\documentclass[lettersize,journal]{IEEEtran}
\usepackage{amsmath,amsfonts}
\usepackage{algorithmic}
\usepackage{algorithm}
\usepackage{array}
\usepackage[caption=false,font=normalsize,labelfont=sf,textfont=sf]{subfig}
\usepackage{textcomp}
\usepackage{stfloats}
\usepackage{url}
\usepackage{verbatim}
\usepackage{graphicx}
\usepackage{cite}

\usepackage{adjustbox}
\usepackage{booktabs}
\usepackage{multirow}
\usepackage{colortbl}
\usepackage{tikz}
\usepackage{xcolor}
\usepackage{soul, color}

\definecolor{bestcolor}{rgb}{ .98,  .706,  .694}
\definecolor{secondcolor}{rgb}{1, .863, .769}
\definecolor{thirdcolor}{rgb}{ 1,  .976,  .89}

\begin{document}

\title{TriDF: Triplane-Accelerated Density Fields for Few-Shot Remote Sensing Novel View Synthesis}

\author{Jiaming Kang, Keyan Chen, Zhengxia Zou, and Zhenwei Shi$^\star$, \\
\vspace{8pt}
Beihang University
}

\maketitle

\begin{abstract}
Remote sensing novel view synthesis (NVS) offers significant potential for 3D interpretation of remote sensing scenes, with important applications in urban planning and environmental monitoring. However, remote sensing scenes frequently lack sufficient multi-view images due to acquisition constraints. While existing NVS methods tend to overfit when processing limited input views, advanced few-shot NVS methods are computationally intensive and perform sub-optimally in remote sensing scenes.
This paper presents TriDF, an efficient hybrid 3D representation for fast remote sensing NVS from as few as 3 input views. 
Our approach decouples color and volume density information, modeling them independently to reduce the computational burden on implicit radiance fields and accelerate reconstruction.
We explore the potential of the triplane representation in few-shot NVS tasks by mapping high-frequency color information onto this compact structure, and the direct optimization of feature planes significantly speeds up convergence. Volume density is modeled as continuous density fields, incorporating reference features from neighboring views through image-based rendering to compensate for limited input data. Additionally, we introduce depth-guided optimization based on point clouds, which effectively mitigates the overfitting problem in few-shot NVS.
Comprehensive experiments across multiple remote sensing scenes demonstrate that our hybrid representation achieves a 30$\times$ speed increase compared to NeRF-based methods, while simultaneously improving rendering quality metrics over advanced few-shot methods (7.4\% increase in PSNR and 3.4\% in SSIM). 
The code is publicly available at  \url{https://github.com/kanehub/TriDF}
\end{abstract}


\section{Introduction}
Novel view synthesis has emerged as a crucial area in computer vision, enabling enhanced 3D understanding through scene reconstruction from multi-view posed images. With the continuous development of remote sensing technology, the demand for 3D interpretation has grown significantly~\cite{10529260, rs14174333, 10654291, liu2024change, 10409216, liu2024rscama}. Novel view synthesis can accurately capture 3D information, including terrain characteristics and building details, which plays a vital role in 3D reconstruction, urban planning, and environmental monitoring~\cite{danilina2018smart, wu2022remote, kang2024few, 9975200, 9934924, 11175207, 10949132}.

\begin{figure}[t]
    \centering
    \includegraphics[width=0.45\textwidth]{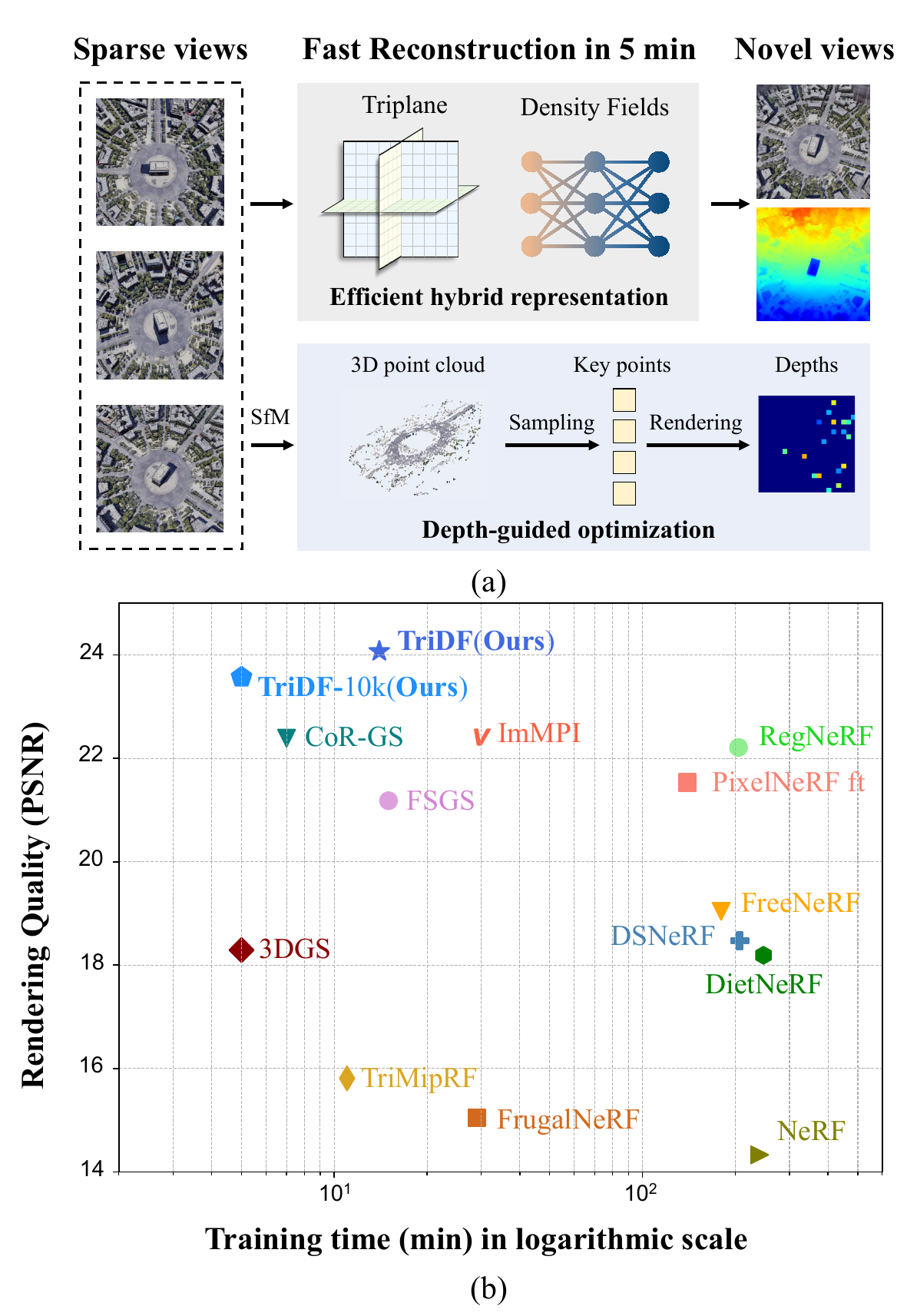}
    \caption{Few-shot remote sensing novel view synthesis. (a) Our hybrid 3D representation TriDF with depth-guided optimization achieves fast reconstruction from only 3 input views. (b) Rendering quality vs. training time on LEVIR-NVS dataset. Our method achieves the best rendering quality compared with advanced few-shot methods, and TriDF-10k (early version with only 10k iterations) is comparable to state-of-the-art 3DGS\cite{kerbl20233d} in reconstruction time.}
    \label{fig:introduction}
\end{figure}

Conventional approaches to scene reconstruction from multi-view images primarily rely on techniques such as structure-from-motion (SfM)~\cite{sturm1996factorization, agarwal2011building, schonberger2016structure}, stereo matching~\cite{scharstein2002taxonomy, hamzah2016literature, brown2003advances}, and multi-view stereo (MVS)~\cite{furukawa2009accurate,lhuillier2005quasi}. However, these methods predominantly focus on point cloud reconstruction and often fail to synthesize high-quality novel view images. Moreover, they exhibit particular sensitivity to challenging conditions frequently encountered in remote sensing scenes, such as poorly textured regions and non-Lambertian surfaces.

Recently, learning-based reconstruction methods~\cite{zbontar2015computing,chang2018pyramid, mayer2016large, yao2018mvsnet} have emerged as a promising alternative, driven by advances in deep learning. Neural Radiance Fields (NeRF)~\cite{mildenhall2020nerf} represents a particularly groundbreaking development in this field~\cite{tewari2020state}. By modeling scenes as continuous implicit functions, NeRF enables the generation of highly realistic novel view images. This breakthrough has inspired numerous NeRF-based approaches, leading to significant advances in various applications, including large-scale reconstruction~\cite{xiangli2022bungeenerf,9879943} and real-time rendering~\cite{muller2022instant, Fridovich-Keil_2022_CVPR}.

Most NVS methods focus on small-scale scenes, individual objects, or synthetic virtual scenes. To successfully reconstruct 3D scenes and render high-quality novel view images, they typically require dense camera arrays to capture hundreds of multi-view images as model inputs. They also rely on long iterative optimizations, leveraging cross-view supervision to ensure spatial consistency. However, for remote sensing scenes, these methods face two critical challenges. The first challenge is data acquisition: Remote sensing images are usually captured by satellites or drones. Due to limitations in flight paths or fuel~\cite{wu2022remote}, it is difficult to obtain hundreds of densely captured multi-view images. As a result, novel view synthesis for remote sensing scenes often suffers from sparse input views, characterized by a limited number of viewpoints and low overlap between views. This sparsity can lead to overfitting on the input images. The second challenge is reconstruction efficiency: The optimization process required for novel view synthesis is time-consuming, which takes dozens of hours to reconstruct a single scene. This makes it unsuitable for real-time applications. To address these challenges, we explore efficient few-shot novel view synthesis in remote sensing scenes, with only 3 input views.

Traditional approaches to few-shot NVS challenges often rely on pretraining with large-scale datasets to extract useful priors~\cite{yu2021pixelnerf, chen2021mvsnerf, 9578424}. However, collecting comprehensive 3D datasets for remote sensing scenes is impractical and inefficient. Alternative methods~\cite{niemeyer2022regnerf, xu2022sinnerf, yang2023freenerf, deng2022depth} incorporate regularization techniques or additional supervision to constrain the reconstruction process, yet these approaches typically perform poorly on remote sensing images. Moreover, NeRF-based methods are hampered by low training efficiency, requiring tens of hours to reconstruct a single scene, which significantly limits their practical applications. Therefore, developing an efficient few-shot novel view synthesis method is crucial for advancing 3D interpretation in remote sensing applications.

In this paper, we propose TriDF, a hybrid 3D representation that achieves both fast training and high-fidelity renderings from sparse views in remote sensing scenes. 
Our motivation stems from the limitations of triplane representations: when supervision is insufficient, triplanes struggle to reconstruct fine-grained geometric structures. Moreover, jointly modeling color and volume density forces the neural radiance fields to learn low-frequency geometry and high-frequency appearance simultaneously, leading to competition for network capacity. This misalignment often results in suboptimal performance and slower convergence. To address these issues, we propose a hybrid explicit–implicit representation that models color and volume density independently.
Specifically, we explore the potential of the triplane representation in few-shot rendering by discretizing the scene space into three mutually orthogonal feature planes.
High-frequency color features are mapped onto this compact structure, accelerating model convergence through direct optimization of explicit feature planes.
Scene geometry is modeled as continuous density fields, while reference features from neighboring views are aggregated using an image-based rendering (IBR) framework to compensate for limited input information.
Since volume density exhibits simpler distribution patterns, this design reduces the optimization burden of the density fields and accelerates the overall training process.

Additionally, we introduce the depth-guided optimization based on 3D point clouds to address overfitting issues in few-shot rendering. Coarse 3D point clouds are generated through Structure-from-Motion (SfM) and Patchmatch stereo techniques, enabling sparse depth supervision to guide accurate scene geometry reconstruction. This approach achieves stable optimization with minimal computational overhead. 
We validate TriDF through extensive experiments on the LEVIR-NVS dataset~\cite{wu2022remote}. As shown in Fig.~\ref{fig:introduction}(b), results demonstrate that our approach significantly enhances few-shot rendering quality across diverse scene types while maintaining a substantial advantage in reconstruction speed. 

Our contributions are summarized as follows:

\begin{itemize}
\item We propose a novel hybrid 3D representation TriDF that incorporates the image-based rendering framework and depth-guided
optimization, which enables fast and few-shot novel view synthesis in remote sensing scenes.

\item We propose to model color and volume density independently. We directly optimize the explicit triplane with high-frequency color information and independently model volume density to reduce the computational burden of density fields, which achieves fast and efficient reconstruction.

\item Extensive experiments on remote sensing scenes demonstrate TriDF's superior performance over various 3D representations and advanced few-shot methods in rendering quality. Moreover, TriDF achieves remarkable training efficiency, completing scene reconstruction in just 5 minutes (30\(\times\) faster than NeRF-based methods) on a single consumer GPU.

\end{itemize}

\section{Related Work}

\subsection{Neural Radiance Fields}

Neural rendering has advanced significantly in the field of novel view synthesis~\cite{yao2018mvsnet, chen2019learning, sitzmann2019scene, niemeyer2020differentiable, chen2023continuous}, with Neural Radiance Fields (NeRF)~\cite{mildenhall2020nerf} emerging as the notable approach. NeRF achieves high-fidelity novel view synthesis by mapping coordinates and view directions to colors and volume densities via implicit neural representations.
 NeRF overcomes the resolution limitations inherent in traditional voxel-based methods by encoding continuous volumetric representations within neural networks.
Inspired by NeRF, subsequent research has enhanced both the quality and efficiency of novel view synthesis while broadening its applications.
NeRF++~\cite{zhang2020nerf++} decomposes the scene space and applies inverted sphere parameterization, extending Neural Radiance Fields to 360° unbounded scenes. Mip-NeRF~\cite{barron2021mip} improves anti-aliasing performance under multi-scale inputs by replacing ray sampling with cone sampling and introducing integrated positional encoding with adaptive frequency adjustment. Instant-NGP~\cite{muller2022instant} achieves significant acceleration while maintaining reconstruction accuracy through the hash encoding of multi-resolution cascades and optimized ray-marching techniques. 

\subsection{Neural Scene Representations}

Neural scene representations can be categorized into explicit, implicit, and hybrid representations~\cite{tewari2020state}. Explicit representations encode scenes using discrete structures with spatial position information, such as point clouds~\cite{xu2022point, kerbl20233d}, meshes~\cite{kulhanek2023tetra}, and voxels~\cite{liu2020neural, Fridovich-Keil_2022_CVPR}.
In contrast, implicit representations model geometry indirectly through continuous functions or neural networks, while hybrid representations leverage both explicit and implicit components to combine their advantages. Each representation presents distinct trade-offs. Explicit representations enable faster evaluation through direct querying of spatial point attributes. However, their memory consumption grows exponentially with increased resolution, limiting their effectiveness for complex scenes. Implicit representations overcome resolution constraints by modeling geometry continuously~\cite{mildenhall2020nerf, 9980428} but suffer from longer training and rendering times due to the computational overhead of neural networks.

NeRF employs a compact implicit representation that requires extensive reconstruction time, significantly limiting its practical applications. Researchers have explored various acceleration methods to address this limitation. KiloNeRF~\cite{reiser2021kilonerf} achieves acceleration by partitioning the scene space into smaller regions and utilizing thousands of tiny Multi-Layer Perceptrons (MLPs) for parallel reconstruction. TermiNeRF~\cite{piala2021terminerf} accelerates training by predicting surface positions and sampling around the surface, thereby reducing the number of samples along each ray.

Other methods focus on utilizing explicit or hybrid scene representations for acceleration. Some researchers employ sparse grids for fast inference~\cite{liu2020neural, yu2021plenoctrees, sun2022direct}. Plenoxels~\cite{Fridovich-Keil_2022_CVPR} extracts a NeRF variant into a sparse voxel grid with spherical harmonics, which enables significantly faster training through direct optimization of explicit representations without neural networks. TensoRF~\cite{chen2022tensorf} models the radiance field as a 4D tensor and factorizes it into low-rank vectors and matrices, reducing both memory usage and reconstruction time.
MINE~\cite{li2021mine} integrates NeRF with Multiplane Images (MPI). Given a single input image, MINE predicts color and volume density at arbitrary depths and renders novel views using homography warping, enabling efficient reconstruction. Due to the inherent sparsity of 3D scenes, volume representations often suffer from redundancy. To address this, ED3G~\cite{chan2022efficient} proposes a compact triplane representation that stores features on axis-aligned planes and employs a lightweight decoder for efficient 3D generation. Similarly, TriMipRF~\cite{hu2023tri} constructs a triplane representation parameterized by three mipmaps, substantially improving training speed while maintaining anti-aliasing effects. 
However, these triplane representations rely on dense input views. In contrast, our proposed method addresses the challenge under sparse-view constraints and achieves efficient reconstruction.
3DGS~\cite{kerbl20233d}, a recently developed reconstruction method, achieves high-quality real-time rendering through direct optimization of 3D Gaussian primitives. However, these methods still rely on dense input images for reconstruction and cannot achieve few-shot rendering.

\subsection{Few-shot Novel View Synthesis}

The vanilla NeRF requires hundreds of input images for optimization, which proves impractical for aerial remote sensing applications. When trained with fewer input images, NeRF tends to overfit the input views, resulting in significant degradation of synthesized images. Consequently, adapting NeRF to sparse inputs has emerged as a critical research direction.

One promising approach involves leveraging prior scene knowledge through pretraining. PixelNeRF~\cite{yu2021pixelnerf} utilizes cross-scene prior knowledge by pretraining on multiple scenes and aggregating local features from neighboring views, enabling reconstruction from a single input image. Similarly, GRF~\cite{trevithick2021grf} combines multi-view geometry with an attention mechanism to reason about reference point visibility. MVSNeRF~\cite{chen2021mvsnerf} employs cost volumes from multi-view stereo geometry for geometry-aware scene inference, facilitating cross-scene generalization. IBRNet~\cite{9578424} utilizes the classical image-based rendering framework and designs a ray transformer to infer sample point attributes. By learning a generic view interpolation function, IBRNet can generalize to novel scenes. However, these methods require extensive pretraining on large multi-view datasets, which are rarely available for remote sensing applications.

Alternative approaches focus on incorporating regularization constraints~\cite{xu2022sinnerf, somraj2023vip, kwak2023geconerf} or additional supervision~\cite{deng2022depth, roessle2022dense, wei2021nerfingmvs}. RegNeRF~\cite{niemeyer2022regnerf} introduces local depth smoothness constraints as regularization to enhance scene structure. DietNeRF~\cite{jain2021putting} employs a pre-trained Vision Transformer (ViT) to enforce semantic consistency across views. FreeNeRF~\cite{yang2023freenerf} implements frequency regularization to anneal positional encoding, enabling few-shot novel view synthesis with minimal computational overhead. DS-NeRF~\cite{deng2022depth} utilizes sparse point clouds for depth supervision to ensure correct scene geometries.
FrugalNeRF~\cite{lin2025frugalnerf} selects pseudo ground truth depth based on the cross-scale reprojection errors to guide its multi-scale weight-sharing voxels in efficiently representing the scene details.
However, methods based on regularization face challenges in balancing quality and efficiency, and some methods require specific designs.

Recently, 3D Gaussian Splatting has gained significant popularity for its remarkable progress in rendering quality and real-time performance. However, it still suffers from a strong dependency on the number of input views. To address the challenge of sparse-view inputs, methods such as FSGS~\cite{zhu2024fsgs} and DNGaussian~\cite{Li_2024_CVPR} introduce monocular depth estimation priors, combined with depth regularization and point cloud densification strategies to enhance the rendering quality. CoR-GS~\cite{zhang2024cor} proposes to jointly train two Gaussian radiance fields, leveraging their mutual inconsistency for collaborative regularization. DropGaussian~\cite{Park_2025_CVPR} introduces a structural regularization technique that randomly removes subsets of Gaussians during training to improve the visibility and update frequency of the remaining ones. However, in remote sensing scenes, it will be difficult to guarantee the reliability of monocular depth estimation. Moreover, the carefully designed regularization strategies employed during Gaussian optimization considerably increase model complexity, thereby restricting their applicability. In contrast, our proposed method achieves efficient and high-quality novel view synthesis from sparse viewpoints without any additional priors.

\subsection{Reconstruction of Remote Sensing Imagery}

Unlike object-level or common datasets, remote sensing imagery provides a distinctive bird’s-eye perspective. Owing to the motion of platforms, multi-view remote sensing images typically exhibit sparse viewpoints and restricted viewing angles. These inherent properties of remote sensing data introduce new challenges for 3D reconstruction and interpretation tasks.

According to the data acquisition source, reconstruction targets can be broadly categorized into satellite remote sensing imagery and aerial remote sensing imagery. Satellite image reconstruction primarily focuses on accurately restoring Digital Surface Models (DSM) or performing surface reconstruction. By integrating the Rational Polynomial Coefficient (RPC) imaging model, such methods handle variations in illumination, shadowing, surface features, and seasonal changes across multi-temporal observations. Representative approaches include S-NeRF~\cite{Derksen_2021_CVPR}, Sat-NeRF~\cite{Mari_2022_CVPR}, and EO-NeRF~\cite{Mari_2023_CVPR}, among others. In contrast, aerial remote sensing imagery, which is typically collected using unmanned aerial vehicles, emphasizes the quality of novel view rendering. ImMPI~\cite{wu2022remote} proposes to reconstruct scenes using a combination of Multi-Plane Image (MPI) representation and implicit neural representation, while large-scale pretraining is employed to obtain 3D scene priors that accelerate the optimization process. It also introduces LEVIR-NVS, a novel-view synthesis dataset for remote sensing. MPNeRF~\cite{10655455} incorporates multi-plane priors to guide NeRF training, exploiting the advantages of MPI to effectively avoid common pitfalls encountered by NeRF in sparse-view aerial scenes. BungeeNeRF~\cite{xiangli2022bungeenerf}, on the other hand, adopts a progressive multi-level supervision framework to achieve multi-scale reconstruction of large-scale urban environments. However, these methods either struggle to handle sparse-view inputs or fall short in terms of reconstruction efficiency and rendering quality.

Furthermore, remote sensing scene reconstruction plays an indispensable role in numerous 3D interpretation tasks. For instance, IRT~\cite{10149540} and R3S~\cite{10950382} utilize neural radiance fields and 3D Gaussian Splatting to represent remote sensing scenes, achieving high-precision and view-consistent semantic segmentation across multiple viewpoints. Therefore, we believe that developing an efficient and high-performance 3D representation in remote sensing scenes is of critical importance for both accurate reconstruction and downstream applications.

\begin{figure*}
    \centering
    \includegraphics[width=0.9\textwidth]{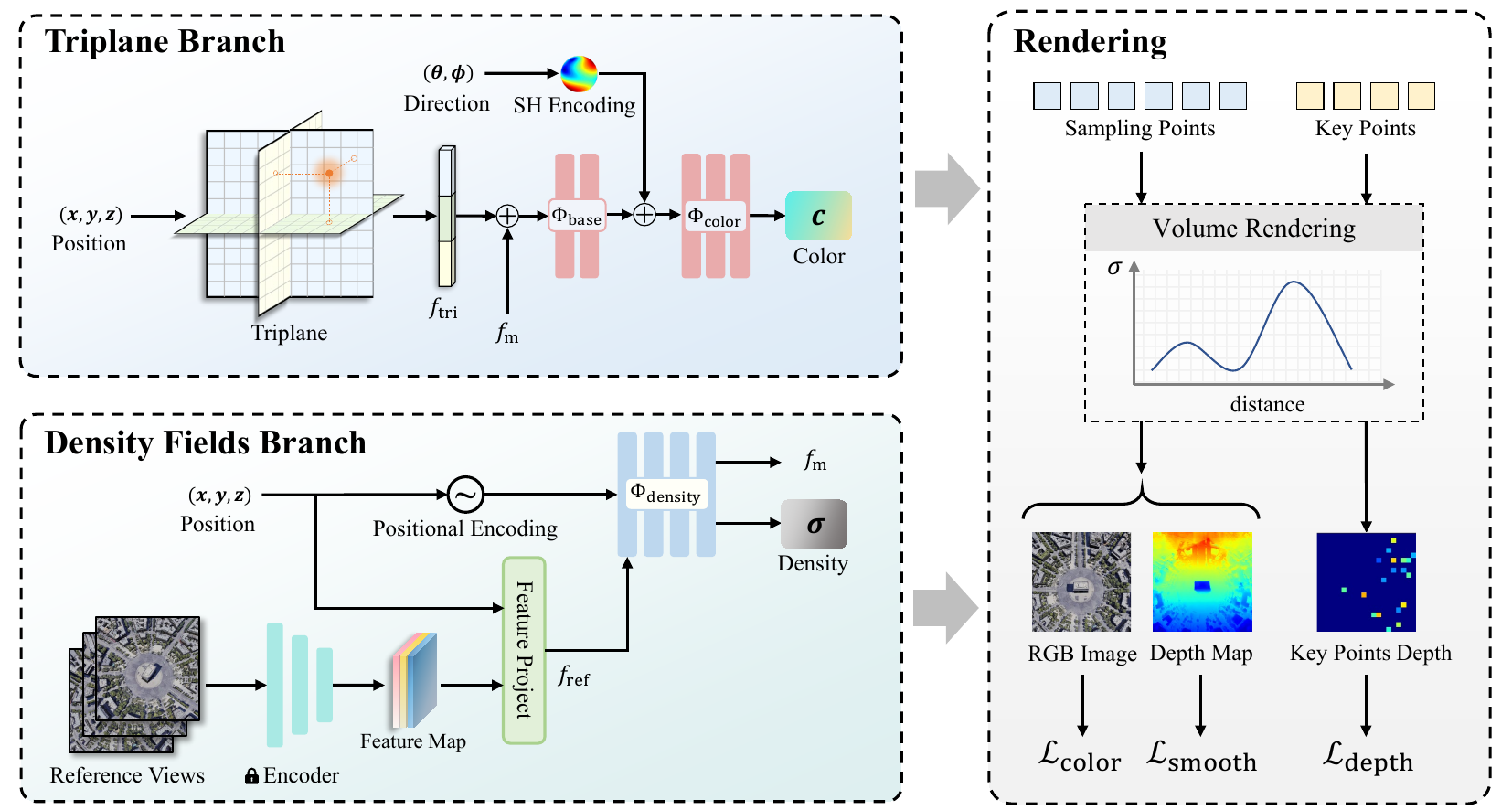}
    \caption{Overview of the proposed TriDF. We introduce an efficient hybrid representation for few-shot novel view synthesis, which takes in sparse posed images and performs fast reconstruction. TriDF consists of a TriPlane branch and a Density Fields branch and separately predicts color and volume density. We also integrate the image-based rendering framework and depth-guided optimization based on 3D point clouds for such hybrid representations to stabilize the training process of few-shot NVS tasks.}
    \label{fig:overview}
\end{figure*}

\section{Method}
\subsection{Overview}

Given a set of sparse multi-view aerial remote sensing images with their corresponding camera poses, we aim to efficiently reconstruct the scene for rendering images from arbitrary novel views. An overview of our proposed method is shown in Fig.~\ref{fig:overview}. To achieve efficient scene reconstruction, we propose a hybrid
3D representation that takes advantage of explicit triplane and implicit density fields, which separately predict the color and volume density of sampling points. High-frequency color features are mapped onto the triplane for direct optimization, and volume density is modeled as continuous density fields through MLPs, which significantly improves training efficiency. Although this hybrid representation accelerates training, its performance drops considerably when input views are sparse. To address this limitation, we incorporate the density fields into an image-based rendering framework to aggregate useful information. Furthermore, we implement a depth-guided optimization approach leveraging point clouds to improve the model’s robustness against sparse inputs while ensuring computational efficiency throughout the training process.

The proposed architecture employs a triplane representation that decomposes space into three orthogonal parameterized planes. The triplane feature vector ${{f}_\mathrm{tri}}$ is generated by projecting each 3D spatial point onto these planes. Subsequently, tiny MLPs map ${{f}_\mathrm{tri}}$ and viewing direction $\mathbf{d} \in \mathbb{R}^{3}$ to determine the point's color $\mathbf{c} \in 
 \mathbb{R}^{3}$. 

The density fields branch uses an MLP to independently predict the volume density of spatial sampling points. To address the challenge of limited information from sparse input views, we implement the image-based rendering framework to aggregate features from neighboring viewpoints. The process begins with feature extraction from input images using a pre-trained CNN model. These features are then projected to obtain corresponding feature vectors for each sampling point. The density fields take both the reference features and spatial coordinates \(\mathbf{x} \in \mathbb{R}^3\) with positional encoding as inputs to predict the volume density $\sigma$.

Given the color and volume density of the sampling points, differentiable volume rendering~\cite{max1995optical} is employed to aggregate the attributes of the sampling points along a ray to obtain the final pixel color. The numerical quadrature form of volume rendering can be formulated as: 

\begin{equation}
\begin{aligned}
 &\hat{C}(\mathbf{r})={\sum_{i=1}^{N}} T_{i}\left(1-\exp \left(-\sigma_{i} \delta_{i}\right)\right) \mathbf{c}_{i}, \\ 
 &T_{i}=\exp \left(-\sum_{j=1}^{i-1} \sigma_{j} \delta_{j}\right),
\end{aligned}
\end{equation}
where \(\mathbf{c}_{i}\) and \(\sigma_{i}\) respectively denote color and volume density,  
 while $\delta_{i}$ represents the distance between adjacent samples. \(N\) refers to the number of sampling points along a ray. 
 \(\hat{C}(\mathbf{r}) \in \mathbb{R}^3\) is the pixel colors and \(T_{i}\) denotes the accumulated transmittance along the ray. The model is optimized through backpropagation using a photometric loss between the predicted and ground truth colors.

A significant limitation of neural radiance fields with sparse input is their tendency to overfit the training views and fail to reconstruct accurate geometries, resulting in blurry synthesized images. To address this limitation, we propose a depth supervision based on point clouds. The approach begins by generating point clouds through Structure-from-Motion (SfM) and Patch Match Stereo (PMS)~\cite{schonberger2016structure,bleyer2011patchmatch}. Through re-projection, we extract key points and their associated depth information from the generated point clouds. These key points function as spatial anchors during the training process, providing additional geometric constraints. This supervision effectively stabilizes the optimization process since the incorporation of depth priors helps maintain geometric consistency across different viewpoints.

\subsection{Triplane-Accelerated Density Fields}

The triplane is an efficient 3D scene representation. It obtains corresponding feature vectors by projecting 3D points onto orthogonal feature planes and retrieves the color and volume density through MLP after feature aggregation. A key advantage of this representation is that it enables direct optimization of explicit triplane features, which allows for the use of lightweight MLP decoders that significantly accelerate reconstruction. Our research aims to extend the triplane representation to novel view synthesis tasks with sparse input while addressing two primary challenges: 1) Inaccurate modeling of volume density: The discrete nature of triplane representation makes it challenging to model fine scene geometry, particularly when limited supervision is provided. 2) Integration difficulties with the image-based rendering framework: The triplane representation struggles to effectively leverage this prior, leading to overly smoothed synthesis results.

To solve these problems, we introduce the density fields that independently model scene geometry while maintaining the triplane representation for color information prediction. This hybrid approach leverages the strengths of both representations while minimizing their respective limitations.

\subsubsection{Triplane Representation}  
The triplane representation decomposes the scene space into three orthogonal feature planes with a fixed resolution, denoted as \( \mathbf{P} = \{\mathbf{P}_{XY}, \mathbf{P}_{YZ}, \mathbf{P}_{ZX}\} \). Each plane consists of a feature map \(\mathbf{P}_{ij} \in \mathbb{R}^{H \times W \times C}\), where \( H \), \( W \), and \( C \) denote the height, width, and channel dimension, respectively. These feature planes are initialized randomly and optimized during the reconstruction process. To obtain the feature vector corresponding to a spatial point, we first project it orthogonally onto the three planes. We then identify the four nearest grid points on each plane and apply bilinear interpolation to compute the feature vectors:

\begin{equation}
    f = \mathrm{TriPlane}\left(\mathbf{x};\mathbf{P}\right).
\end{equation}

The final feature vector \(f_\mathrm{tri}\) is obtained by concatenating the feature vectors from the three planes: \(f_\mathrm{tri}=\mathrm{concat}({f}_{XY},{f}_{YZ},{f}_{ZX})\).

To process these features, we employ a compact MLP that combines the triplane features \({f}_\mathrm{tri}\) and density features \({f}_\mathrm{m}\) to generate intermediate features. Subsequently, we encode view directions on the spherical harmonics basis \(\mathrm{SH}(\cdot)\). The encoded directions and intermediate features \({f}_\mathrm{base}\) are then processed through the Color MLP to predict the colors. This pipeline can be formally expressed as:

\begin{equation}
\begin{aligned}
    {f}_\mathrm{base} &= {\Phi_\mathrm{base}\left({f}_\mathrm{tri},~{f}_\mathrm{m}\right)},\\
    {c} &= {\Phi_\mathrm{color}\left({f}_\mathrm{base}, \mathrm{SH}(\mathbf{d})\right)}.
\end{aligned} 
\end{equation}

The direct optimization of triplane features allows us to utilize a tiny MLP, resulting in accelerated model convergence during reconstruction.

\subsubsection{Density Fields}  
We model the volume density of spatial points using the density fields, aggregating reference features from neighboring views to enhance input information. Initially, we employ a pre-trained encoder to extract pixel-aligned features from each reference image:

\begin{equation}
    \mathbf{F}_k=\mathrm{\Phi_{encoder}}(\mathbf{I}_k),  k=1,…,M ,
\end{equation}
where \(\mathbf{I}_k \in \mathbb{R}^{3\times H \times W}\) denotes the reference image  and \(\mathbf{F}_k \in \mathbb{R}^{C_r \times H_r \times W_r}\) represents the corresponding feature map. \(M\) denotes the number of reference images. Subsequently, we project each spatial point \(\mathbf{x}\) onto the reference images:

\begin{equation}
    \pi_k \left(\mathbf{x}\right)= \mathbf{K}_k \mathbf{E}_k \mathbf{x},
\end{equation}
where \( \mathbf{K}_k \in \mathbb{R}^{3 \times 3}\) denotes the intrinsic matrix and  \( \mathbf{E}_k \in \mathbb{R}^{3 \times 4}\) represents world-to-camera extrinsic matrix of the \(k\)-th reference image, respectively. Using the projected positions, we obtain feature vectors through bilinear interpolation from the reference images and concatenate them:

\begin{equation}
\begin{aligned}
    {f}_k &= \mathbf{F}_k\left(\pi_k\left(\mathbf{x}\right)\right) \\
    {f}_\mathrm{ref} &= \mathrm{concat}\left({f}_k\right).
\end{aligned}  
\end{equation}

The density fields are then formulated as:

\begin{equation}
    \sigma,~{f}_{\mathrm{m}} = \mathrm{\Phi_{density}}\left(\gamma\left(\mathbf{x}\right), {{f}_\mathrm{ref}}\right),
\end{equation}
where \(\gamma(\cdot)\) represents positional encoding and \({f}_\mathrm{m}\) denotes density features. Due to spatial continuity, the volume density distribution along a ray is relatively simple compared to color, which often contains complex high-frequency components such as intricate textures. Therefore, modeling volume density alone is more efficient than simultaneously predicting both color and density, which leads to faster convergence.

The hybrid representation is optimized by minimizing the per-pixel mean squared error between the rendered color and the ground truth:
\begin{equation}
   \mathcal{L}_{\mathrm{color}}=\mathbb{E}_{\mathbf{r}\in \mathcal{R}_i}\|\hat{{C}}(\mathbf{r})-{C}(\mathbf{r})\|_2^2 ,
\end{equation}

\noindent where \( \mathcal{R}_i\) indicates the set of sampled rays.

\subsection{Depth-Guided Optimization with 3D Point Clouds}

Neural radiance fields learn the 3D structure of a scene by evaluating multi-view consistency across 2D images. However, with limited input images, models often overfit the training views, resulting in inaccurate geometry reconstruction and blurry image synthesis. To address this challenge, RegNeRF~\cite{niemeyer2022regnerf} introduces a sampling space annealing method that incorporates an inductive bias about the scene distribution. This approach initially restricts the sampling space to a small region and gradually expands it during training, effectively mitigating overfitting. Similarly, FreeNeRF~\cite{yang2023freenerf} adopts a frequency regularization technique that progressively increases positional encoding frequencies throughout training. This strategy enables the model to learn hierarchically, from coarse structures at low frequencies to fine details at high frequencies, thereby reducing noise and artifacts. Although these methods have demonstrated considerable success, they are primarily limited to implicit representations and show suboptimal performance with hybrid representations. Thus, we aim to develop a robust optimization strategy specifically designed for hybrid representations in few-shot novel view synthesis. 

 We leverage point clouds estimated from Structure-from-Motion (SfM) as depth guidance to ensure accurate geometry reconstruction without overfitting to the training views. Since most reconstruction processes require SFM to estimate camera poses, this approach introduces no additional computational overhead. As discussed in RegNeRF~\cite{niemeyer2022regnerf}, overfitting manifests as excessive volume density values near the ray origin, making depth maps too close to the camera plane. To address this, we anchor the depth map using point cloud positions to reconstruct accurate geometry.

Due to the limited number of training images, SfM-estimated point clouds are sparse, containing only approximately \(10^2\) points, which provides insufficient depth guidance. To overcome this limitation, we employ PatchMatch Stereo~\cite{bleyer2011patchmatch} for additional stereo matching and point cloud fusion, generating dense 3D point clouds with approximately \(10^3\) points. These processes are implemented using COLMAP with minimal computational cost.

Given a 3D point, we project it onto all 2D views to obtain keypoints and their corresponding depths. First, we transform the 3D point from world coordinates to camera coordinates:

\begin{equation}
\begin{bmatrix}X_c\\Y_c\\Z_c\\1\end{bmatrix}=\mathbf{EP}_w=\begin{bmatrix}\mathbf{R}&\mathbf{T}\\0&1\end{bmatrix}\begin{bmatrix}X_w\\Y_w\\Z_w\\1\end{bmatrix}.
\end{equation}
where \(\mathbf{R} \in \mathbb{R}^{3 \times 3}\) and \(\mathbf{T} \in \mathbb{R}^{3 \times 1}\) denote rotation matrix and translation matrix.

Next, we reproject the 3D point onto the image plane to obtain the pixel coordinates \((u,v)\), where \( Z_c \) represents the corresponding depth value and \(\mathbf{K} \in \mathbb{R}^{3 \times 3}\) denotes the intrinsic matrix, 
\begin{equation}
Z_c\begin{bmatrix}u\\v\\1\end{bmatrix}=\mathbf{KP}_c=\begin{bmatrix}f_x&0&c_x\\0&f_y&c_y\\0&0&1\end{bmatrix}\begin{bmatrix}X_c\\Y_c\\Z_c\end{bmatrix}.
\end{equation}

For the projected 2D key points obtained from 3D point clouds, we perform additional volume rendering to obtain depth information in each iteration:
\begin{equation}
    \hat{d}=\sum_{i=1}^NT_i(1-\exp(-\sigma_i\delta_i))t_i,
\end{equation}
where \(t_i\) denotes the position of the sampling point.

Due to inherent inaccuracies in estimated 3D point clouds, we compute an adaptive weight for each point based on multi-view consistency. These inaccuracies primarily arise from two sources: inaccurate spatial positions of point clouds and projection ambiguities at edge regions. Points with spatial inaccuracies exhibit significant color discrepancies across multi-view projections. Besides, points near edge regions are highly sensitive to positional errors, where minor positional shifts will result in pronounced variations in both depth and color. We quantify these errors by measuring color differences between the 3D point cloud and its 2D projections. The color error between pixel \(i\) and \(j\) is defined as:

\begin{equation}
    S(\mathbf{c}_i,\mathbf{c}_j) = \frac{1}{3}\|\mathbf{c}_i-\mathbf{c}_j\|_1 .
\end{equation}

We compute both the standard deviation of colors between projected points and the color error between the point cloud and its projections:

\begin{equation}
\begin{aligned}
    e_k^{i}&=\left(\frac{1}{M-1}\sum_{j=1}^{M}S\left(\mathbf{c}_j^i,\overline{\mathbf{c}}^i\right)\right)^{\frac{1}{2}} + S(\mathbf{c}_k^i,\mathbf{p}^i)
\end{aligned}
\end{equation}
where \(e_k^{i}\) represents the error of the keypoint projected by the $i$-th point cloud on the $k$-th view, $M$ denotes the number of reference views, \(c_j^{i}\) represents the color of the $i$-th point cloud projected on the $j$-th view, \(\overline{\mathbf{c}}^i\) represents the average color of all projected keypoints, and \(\mathbf{p}^i\) is the color of the point cloud.

Then we calculate the adaptive weight for each key point:

\begin{equation}
\begin{aligned}
     \alpha_k^{i}&=\left(1-e_{k}^{i}\right)^{2},\\ w_k^{i}&=\mathrm{clamp}(\alpha_k^{i},0,1),   
\end{aligned}
\end{equation}
where \(w^i_k\) represents the weight for keypoint projected by the $i$-th point cloud on the $k$-th view.

The depth loss is then computed as the weighted mean square error between the predicted depth and the point cloud depth:

\begin{equation}
\mathcal{L}_{\mathrm{depth}}=\frac{1}{N_s}\sum_{i=1}^{N_s}w_i(\hat{d_{i}} - d_{i})^2,
\end{equation}
where \(\hat{d_i}\) and \({d_i}\) denote the predicted depth and the depth of point clouds, respectively, and \(N_s\) refers to the number of key points sampled in each iteration.

We employ adaptive weights to adjust the influence of different keypoints, assigning lower weights to those associated with higher errors. However, this strategy does not fully mitigate the impact of inaccuracy. Consequently, such depth information can only serve as a guidance signal rather than ground truth. In later optimization stages, these errors may compromise the scene's geometric structure and reduce training efficiency. Therefore, we apply depth guidance only during the early training stages. Subsequently, we refine the model using a combination of photometric consistency loss and additional regularization strategies.

\subsection{Optimization}
To enhance the geometric structure of the scene and mitigate rendering artifacts caused by high-frequency noise while preserving original details, we implement edge-aware smoothness regularization in the latter stages of training. During each iteration, we render additional patches from both training and test viewpoints. The regularization employs first-order gradients of RGB values to promote depth map smoothness in flat regions:

\begin{equation}
   \mathcal{L}_{\mathrm{smooth}}=|\partial_xd_t^*|e^{-|\partial_xI_t|}+|\partial_yd_t^*|e^{-|\partial_yI_t|},
\end{equation}
where \(d_t^*\) represents the disparity map and \(I_t\) denotes the RGB value of the rendered patches. The overall loss function is defined as:

\begin{equation}
    \mathrm{{\mathcal L}_{total}}=\mathrm{{\mathcal L}_{color}}+\lambda_{1}\mathrm{{\mathcal L}_{depth}}+\lambda_{2}\mathrm{{\mathcal L}_{smooth}},
\end{equation}
where \(\lambda_1\) and \(\lambda_2\) are weighting factors. In our implementation, we initially set \(\lambda_1=0.001\) and \(\lambda_2=0\) to enable depth-guided optimization during the early training phase. As training progresses, we modify these parameters, setting \(\lambda_1=0\) to improve computational efficiency and \(\lambda_2=1.0\) to activate depth smoothness regularization.

\section{Experiments}
\subsection{Dataset and Experimental Setup}

\subsubsection{LEVIR-NVS\cite{wu2022remote}}  
The LEVIR-NVS dataset comprises 16 diverse remote sensing scenes specifically designed for novel view synthesis tasks. These scenes contain various environmental types, including urban areas, villages, mountainous regions, and buildings. Several scenes have complex structures accompanied by detailed backgrounds. The image resolution is $512\times512$. We uniformly sampled three viewpoints per scene for training, and the remaining are used for evaluation across all methods.

\subsubsection{Baselines} 
We evaluate TriDF against various 3D scene representations, including implicit representations like NeRF \cite{mildenhall2020nerf}, hybrid representations TriMipRF \cite{hu2023tri} and ImMPI \cite{wu2022remote}, and explicit representations 3DGS \cite{kerbl20233d}. We also compare with state-of-the-art few-shot novel view synthesis methods: DietNeRF \cite{jain2021putting}, DSNeRF \cite{deng2022depth}, PixelNeRF \cite{yu2021pixelnerf}, RegNeRF \cite{niemeyer2022regnerf}, FreeNeRF \cite{yang2023freenerf}, and FrugalNeRF \cite{lin2025frugalnerf}. Besides, we compare with advanced 3DGS few-shot methods, including FSGS \cite{zhu2024fsgs} and CoR-GS \cite{zhang2024cor}. For a fair comparison, we generate initial point clouds from training view images for 3DGS-based methods and fine-tune PixelNeRF on each scene. Similar to previous novel view synthesis approaches, we report the average PSNR, SSIM, and LPIPS for all methods. To evaluate model efficiency, we also report the average training time per scene for each model and the rendering speed.

\begin{figure*}
    \centering
    \includegraphics[width=\textwidth]{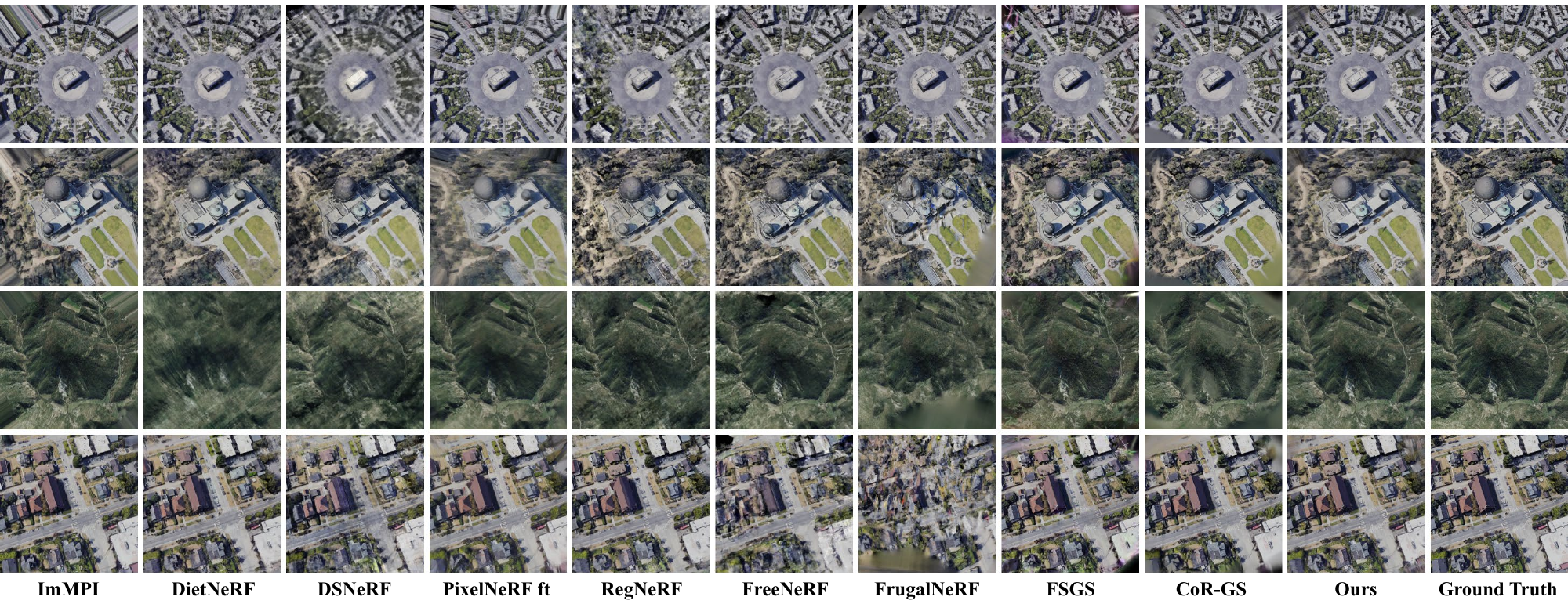}
    \caption{Qualitative comparison of test-view rendering images with 3 input views on LEVIR-NVS dataset.}
    \label{fig:vis_result}
\end{figure*}

\subsection{Implementation Details} 
The proposed hybrid representation focuses on novel view synthesis for remote sensing scenes with sparse inputs. Our approach decouples the modeling of color and volume density. 
 Specifically, the color information is encoded through a triplane representation, while the volume density is independently modeled via the continuous density fields. Additionally, we employ the image-based rendering framework to fully utilize the input image information.

\subsubsection{Architecture Details}
 The triplane branch independently predicts color information by sampling from feature planes. We employ fixed-resolution feature planes (512 \(\times\)  512) with 8 channels per plane. The sampled triplane features are concatenated with predicted density features before being processed by the base MLP to generate intermediate features. These features are combined with direction vectors (encoded using 4th-order spherical harmonics) and processed through a tiny color MLP to predict final RGB values. Both MLPs utilize the fully fused MLP structure for efficiency, with a width of 128 and depths of 2 and 4 for the base MLP and the color MLP, respectively.

 We integrate the density fields with an image-based rendering framework for volume density prediction. A pre-trained CNN extracts per-pixel features from input images, constructing a feature pyramid by concatenating upsampled intermediate feature maps from each layer on the channel dimension. The reference image feature maps maintain a dimension of 64. The system concatenates all projected feature vectors with positionally encoded coordinates (frequency = 6) before processing them through the density MLP. The density MLP is implemented using CUTLASS architecture with a width of 512 and a depth of 8.

\begin{table}[]
    \centering
    \caption[toc entry]{Quantitative comparison of training efficiency and rendering quality between different methods. 
     The best, second-best, and third-best entries are marked in     
   \begin{tikzpicture}
        \draw[fill=bestcolor, draw=white] (0,0) rectangle (0.6,0.25);
    \end{tikzpicture}, 
    \begin{tikzpicture}
        \draw[fill=secondcolor, draw=white] (0,0) rectangle (0.6,0.25);
    \end{tikzpicture}, and 
    \begin{tikzpicture}
        \draw[fill=thirdcolor, draw=white] (0,0) rectangle (0.6,0.25);
    \end{tikzpicture}, 
    respectively.
    }
     \setlength{\extrarowheight}{1pt}
    \resizebox{0.5\textwidth}{!}{
    \begin{tabular}{l|ccc|c|c}
    \toprule
        Method  & PSNR$\uparrow$ & SSIM$\uparrow$ & LPIPS$\downarrow$ & Time$\downarrow$ & FPS$\uparrow$ \\
        
    \midrule
    NeRF~\cite{mildenhall2020nerf} & 14.33 & 0.172 & 0.633 & 4.0 h & 0.18 \\
    TriMipRF~\cite{hu2023tri}   & 15.81 & 0.263 & 0.597 & 11 min & 1.22 \\
    3DGS~\cite{kerbl20233d} & 18.29 & 0.593 & 0.313 & 5 min & 280 \\
    ImMPI~\cite{wu2022remote} & \cellcolor{thirdcolor}22.41 & 0.731 & \cellcolor{thirdcolor}0.262 & 30 min & 43 \\
    \midrule
    DietNeRF~\cite{jain2021putting}& 18.19 & 0.420 & 0.466 & 4.1 h & 0.14 \\
    DSNeRF~\cite{deng2022depth} & 18.47 & 0.496 & 0.482 & 3.4 h & 0.17 \\
    PixelNeRF~ft~\cite{yu2021pixelnerf}& 21.17 & 0.646 & 0.375 & 2.3 h & 0.04 \\
    RegNeRF~\cite{niemeyer2022regnerf}& 19.83 & 0.695 & 0.389 & 3.4 h & 0.19 \\
    FreeNeRF~\cite{yang2023freenerf}& 19.04 & 0.524 & 0.373 & 3.0 h & 0.19 \\
    FrugalNeRF~\cite{lin2025frugalnerf} & 15.05 & 0.304 & 0.483 & 29 min & 0.25 \\
    FSGS~\cite{zhu2024fsgs}& 21.18 & 0.772 & 0.230 & 15 min & 343 \\
    CoR-GS~\cite{zhang2024cor}& 22.39 & \cellcolor{secondcolor}0.793 & \cellcolor{bestcolor}0.212 & 7 min & 268 \\
    \midrule
    TriDF-10K & \cellcolor{secondcolor}23.57 & \cellcolor{thirdcolor}0.787 & 0.275 & 5 min &  0.20 \\
    \textbf{TriDF (Ours)}& \cellcolor{bestcolor}24.07 & \cellcolor{bestcolor}0.820 & \cellcolor{secondcolor}0.213 & 14 min & 0.20 \\    
    \bottomrule
    \end{tabular}
    }
    \label{tab:result_comparison}
\end{table}

\subsubsection{Training Details}
We implement TriDF using the PyTorch framework, with initial 3D point clouds estimated from training views and poses using COLMAP. Our implementation employs a batch size of 1024 pixel rays with 128 coarse samples per ray, and utilizes Nerfacc \cite{li2022nerfacc} for efficient empty space skipping. The optimization process consists of 30k total iterations: the first 10k iterations incorporate depth guidance optimization, followed by 20k iterations with edge-aware smoothness loss for geometric constraint. For each iteration, we process 64 key points and randomly render a $16 \times 16 $ patch (stride = 4) for the edge-aware smoothness loss computation. The model is optimized using AdamW with an initial learning rate of 0.0001. All experiments are conducted on a single Nvidia GeForce RTX 4090 GPU.

\begin{table*}[htbp]
  \centering
  \caption{Quantitative comparison per scene on LEVIR-NVS dataset.
  }
    \begin{adjustbox}{width=\linewidth}
    \begin{tabular}{c|c|ccccccccccccccccc}
    \toprule
    \toprule
    \multicolumn{1}{c}{\rotatebox{60}{Metrics}} & \multicolumn{1}{c}{\rotatebox{60}{Methods}} & \rotatebox{60}{Building\#1} & \rotatebox{60}{Church} & \rotatebox{60}{College} & \rotatebox{60}{Mountain\#1} & \rotatebox{60}{Mountain\#2} & \rotatebox{60}{Observation} & \rotatebox{60}{Building\#2} & \rotatebox{60}{Town\#1} & \rotatebox{60}{Stadium} & \rotatebox{60}{Town\#2} & \rotatebox{60}{Mountain\#3} & \rotatebox{60}{Town\#3} & \rotatebox{60}{Factory} & \rotatebox{60}{Park} & \rotatebox{60}{School} & \rotatebox{60}{Downtown} & \rotatebox{60}{Mean} \\
    \midrule
    \midrule
    \multirow{11}[0]{*}{PSNR}   & NeRF\cite{mildenhall2020nerf}  & 12.79  & 10.96  & 12.93  & 18.26  & 17.13  & 14.99  & 12.92  & 13.82  & 15.34  & 12.58  & 21.03  & 12.89  & 14.02  & 14.48  & 13.27  & 11.91  & 14.33 \\
          & TrimipRF\cite{hu2023tri}  & 14.19  & 12.01  & 14.11  & 17.99  & 18.12  & 14.40  & 13.53  & 15.58  & 17.03  & 14.40  & 19.12  & 14.05  & 20.74  & 18.57  & 15.02  & 14.05  & 15.81 \\
          & 3DGS\cite{kerbl20233d}  & 16.27  & 15.16  & 16.27  & 20.16  & 21.58  & 16.68  & 16.73  & 18.54  & 18.46  & 17.22  & 24.21  & 17.93  & 20.25  & 20.64  & 16.49  & 16.04  & 18.29 \\
          & ImMPI\cite{wu2022remote}  & \cellcolor{thirdcolor}18.75  & 17.27  & \cellcolor{secondcolor}20.99  & \cellcolor{secondcolor}26.48  & \cellcolor{thirdcolor}25.54  & 19.55  & \cellcolor{thirdcolor}18.65  & \cellcolor{thirdcolor}22.35  & 22.98  & \cellcolor{secondcolor}22.22  & \cellcolor{secondcolor}29.74  & \cellcolor{secondcolor}22.08  & \cellcolor{secondcolor}24.84  & \cellcolor{thirdcolor}24.81  & \cellcolor{secondcolor}22.17  & \cellcolor{thirdcolor}20.20  & \cellcolor{secondcolor}22.41 \\
          & DietNeRF\cite{jain2021putting}  & 16.67  & 11.46  & 13.66  & 18.78  & 18.14  & 17.93  & 14.33  & 13.87  & 21.95  & \cellcolor{thirdcolor}22.03  & 25.42  & 14.90  & \cellcolor{thirdcolor}24.25  & 24.52  & 20.66  & 12.42  & 18.19 \\
          & DSNeRF\cite{deng2022depth}  & 16.56  & 13.46  & 16.06  & 20.52  & 20.87  & 16.05  & 15.18  & 18.28  & 17.47  & 19.32  & 27.28  & 16.10  & 21.89  & 23.08  & 18.71  & 14.64  & 18.47 \\
          & PixelNeRF ft\cite{yu2021pixelnerf}  & \cellcolor{secondcolor}19.00  & 16.25  & 19.18  & 22.91  & 25.38  & 18.62  & 17.83  & 21.69  & 22.72  & 21.81  & 24.97  & 20.33  & 23.01  & \cellcolor{secondcolor}26.38  & 19.79  & 18.92  & 21.17 \\
          & RegNeRF\cite{niemeyer2022regnerf}  & 13.52  & 14.74  & 17.63  & 20.88  & 20.57  & 15.24  & 14.81  & 20.84  & 22.35  & 21.15  & 26.50  & 21.20  & 23.49  & 24.74  & 20.11  & 19.46  & 19.83 \\
          & FreeNeRF\cite{yang2023freenerf}  & 15.84  & 15.99  & 19.74  & 19.64  & 21.41  & 13.76  & 15.85  & 20.58  & 23.41  & 14.66  & 20.56  & 19.05  & 21.03  & 23.84  & \cellcolor{thirdcolor}21.81  & 17.46  & 19.04 \\
          & FrugalNeRF~\cite{lin2025frugalnerf}  & 18.65  & 10.84  & 14.04  & 18.01  & 19.11  & 12.52  & 11.78  & 13.30  & 15.73  & 12.20  & 29.30  & 12.52  & 12.51  & 17.35  & 12.12  & 11.46  & 15.09 \\
          & FSGS\cite{zhu2024fsgs}  & 17.69  & \cellcolor{thirdcolor}17.52  & 20.01  & 25.03  & 24.39  & \cellcolor{thirdcolor}19.76  & 18.36  & 20.74  & \cellcolor{thirdcolor}23.42  & 19.92  & 26.46  & 21.11  & 22.47  & 23.93  & 18.42  & 19.63  & 21.18 \\
          & CoR-GS\cite{zhang2024cor}  & 18.39  & \cellcolor{secondcolor}18.13  & \cellcolor{thirdcolor}20.57  & \cellcolor{thirdcolor}26.30  & \cellcolor{secondcolor}25.68  & \cellcolor{secondcolor}20.44  & \cellcolor{secondcolor}20.09  & \cellcolor{secondcolor}22.64  & \cellcolor{secondcolor}23.83  & 21.58  & \cellcolor{thirdcolor}29.41  & \cellcolor{thirdcolor}21.26  & 23.72  & 24.37  & 21.32  & \cellcolor{secondcolor}20.49  & \cellcolor{thirdcolor}22.39 \\
          & Ours  & \cellcolor{bestcolor}21.42  & \cellcolor{bestcolor}19.56  & \cellcolor{bestcolor}22.98  & \cellcolor{bestcolor}27.70  & \cellcolor{bestcolor}27.05  & \cellcolor{bestcolor}20.70  & \cellcolor{bestcolor}20.74  & \cellcolor{bestcolor}24.47  & \cellcolor{bestcolor}25.45  & \cellcolor{bestcolor}22.36  & \cellcolor{bestcolor}30.70  & \cellcolor{bestcolor}22.86  & \cellcolor{bestcolor}25.28  & \cellcolor{bestcolor}27.62  & \cellcolor{bestcolor}24.02  & \cellcolor{bestcolor}22.19  & \cellcolor{bestcolor}24.07 \\

    \midrule
    \multirow{12}[0]{*}{SSIM} & NeRF\cite{mildenhall2020nerf} & 0.139& 	0.079& 	0.138& 	0.279& 	0.199& 	0.236& 	0.117& 	0.114& 	0.197& 	0.122& 	0.351& 	0.151& 	0.208& 	0.144& 	0.180& 	0.092& 	0.172  \\
          & TrimipRF\cite{hu2023tri} & 0.218 &	0.112 &	0.172 &	0.263 &	0.238 &	0.151 &	0.137 &	0.236 &	0.323 &	0.253 &	0.256 &	0.210 &	0.672 &	0.498 &	0.272 &	0.202 &	0.263  \\
          & 3DGS\cite{kerbl20233d} & 0.622 &	0.460 &	0.385 &	0.447 &	0.687 &	0.479 &	0.521 &	0.626 &	0.646 &	0.617 &	0.694 &	0.606 &	0.857 &	0.846 &	0.478 &	0.524 &	0.593 \\
          & ImMPI\cite{wu2022remote} & 0.697 &	0.580 &	0.676 &	0.765 &	0.750 &	0.576 &	0.631 &	0.750 &	0.796 &	0.767 &	0.814 &	0.740 &	0.855 &	0.841 &	0.745 &	0.707 &	0.731  \\
          & DietNeRF\cite{jain2021putting} & 0.551 &	0.091 &	0.154 &	0.321 &	0.235 &	0.444 &	0.232 &	0.126 &	0.657 &	0.742 &	0.550 &	0.292 &	0.808 &	0.785 &	0.612 &	0.116 &	0.420  \\ 
          & DSNeRF\cite{deng2022depth} & 0.446& 	0.337& 	0.390& 	0.442& 	0.533& 	0.373& 	0.326& 	0.503& 	0.484& 	0.623& 	0.669& 	0.471& 	0.653& 	0.760& 	0.569& 	0.350& 	0.496   \\
          & PixelNeRF ft\cite{yu2021pixelnerf} & \cellcolor{thirdcolor}0.725& 	0.424& 	0.522& 	0.523& 	0.690& 	0.457& 	0.521& 	0.684& 	0.764& 	0.788& 	0.531& 	0.646& 	0.878& 	\cellcolor{secondcolor}0.889& 	0.700& 	0.590& 	0.646 \\
          & RegNeRF\cite{niemeyer2022regnerf} & 0.390& 	0.495& 	0.580& 	0.680& 	0.705& 	0.457& 	0.410& 	0.756& 	\cellcolor{thirdcolor}0.844& 	\cellcolor{secondcolor}0.836& 	0.851& 	0.781& 	\cellcolor{secondcolor}0.897& 	\cellcolor{thirdcolor}0.884& 	\cellcolor{secondcolor}0.782& 	\cellcolor{thirdcolor}0.745& 	0.695 \\
          & FreeNeRF\cite{yang2023freenerf} & 0.441 &	0.482 &	0.541 &	0.414 &	0.511 &	0.140 &	0.378 &	0.627 &	0.788 &	0.331 &	0.294 &	0.643 &	0.714 &	0.810 &	0.692 &	0.584 &	0.524  \\
          & FrugalNeRF~\cite{lin2025frugalnerf} & \cellcolor{secondcolor}0.749 & 0.105 & 0.245 & 0.313 & 0.398 & 0.138 & 0.136 & 0.147 & 0.341 & 0.185 & \cellcolor{bestcolor}0.892 & 0.184 & 0.224 & 0.511 & 0.197 & 0.128 & 0.306 \\
          & FSGS~\cite{zhu2024fsgs} & 0.696 & \cellcolor{thirdcolor}0.642 & \cellcolor{thirdcolor}0.682 & \cellcolor{secondcolor}0.813 & \cellcolor{secondcolor}0.840 & \cellcolor{secondcolor}0.702 & \cellcolor{thirdcolor}0.703 & \cellcolor{thirdcolor}0.795 & 0.837 & 0.798 & 0.864 & \cellcolor{thirdcolor}0.806 & 0.870 & 0.863 & 0.708 & 0.732 & \cellcolor{thirdcolor}0.772 \\
          & CoR-GS~\cite{zhang2024cor} & 0.707 & \cellcolor{secondcolor}0.665 & \cellcolor{secondcolor}0.693 & \cellcolor{bestcolor}0.825 & \cellcolor{bestcolor}0.843 & \cellcolor{bestcolor}0.705 & \cellcolor{secondcolor}0.749 & \cellcolor{secondcolor}0.810 & \cellcolor{secondcolor}0.857 & \cellcolor{thirdcolor}0.824 & \cellcolor{thirdcolor}0.883 & \cellcolor{secondcolor}0.816 & \cellcolor{thirdcolor}0.885 & 0.860 & \cellcolor{thirdcolor}0.781 & \cellcolor{secondcolor}0.792 & \cellcolor{secondcolor}0.793 \\
          & Ours  & 
          \cellcolor{bestcolor}0.825 & \cellcolor{bestcolor}0.693 & \cellcolor{bestcolor}0.759 & \cellcolor{thirdcolor}0.796 & \cellcolor{thirdcolor}0.839 & \cellcolor{thirdcolor}0.660 & \cellcolor{bestcolor}0.756 & \cellcolor{bestcolor}0.847 & \cellcolor{bestcolor}0.882 & \cellcolor{bestcolor}0.848 & \cellcolor{secondcolor}0.886 & \cellcolor{bestcolor}0.822 & \cellcolor{bestcolor}0.913 & \cellcolor{bestcolor}0.918 & \cellcolor{bestcolor}0.857 & \cellcolor{bestcolor}0.813 & \cellcolor{bestcolor}0.820   \\

    \midrule
    \multirow{12}[0]{*}{LPIPS} & NeRF\cite{mildenhall2020nerf} & 0.636& 	0.678& 	0.677& 	0.592& 	0.624& 	0.571& 	0.652& 	0.649& 	0.632& 	0.659& 	0.544& 	0.668& 	0.588& 	0.647& 	0.651& 	0.658& 	0.633 \\
          & TrimipRF\cite{hu2023tri} & 0.624 &	0.656 &	0.664 &	0.596 &	0.596 &	0.636 &	0.652 &	0.631 &	0.570 &	0.608 &	0.574 &	0.656 &	0.361 &	0.506 &	0.602 &	0.626 &	0.597   \\
          & 3DGS\cite{kerbl20233d} & 0.282 &	0.393 &	0.450 &	0.433 &	0.267 &	0.365 &	0.359 &	0.281 &	0.277 &	0.296 &	0.287 &	0.317 &	\cellcolor{secondcolor}0.118 &	\cellcolor{secondcolor}0.131 &	0.397 &	0.363 &	0.313  \\
          & ImMPI\cite{wu2022remote} & 0.284 &	0.341 &	0.312 &	0.281 &	0.279 &	0.352 &	0.311 &	0.251 &	0.196 &	0.232 &	0.241 &	0.259 &	0.157 &	0.179 &	\cellcolor{thirdcolor}0.228 &	0.295 &	0.262   \\
          & DietNeRF\cite{jain2021putting} & 0.397 &	0.627 &	0.609 &	0.563 &	0.578 &	0.431 &	0.569 &	0.597 &	0.322 &	0.261 &	0.464 &	0.556 &	0.233 &	0.264 &	0.371 &	0.608 &	0.466  \\
          & DSNeRF\cite{deng2022depth} &  0.585& 	0.528& 	0.525& 	0.529& 	0.472& 	0.523& 	0.582& 	0.451& 	0.454& 	0.379& 	0.436& 	0.472& 	0.468& 	0.333& 	0.408& 	0.568& 	0.482  \\
          & PixelNeRF ft\cite{yu2021pixelnerf}& 0.298& 	0.508& 	0.476& 	0.495& 	0.356& 	0.483& 	0.465& 	0.355& 	0.269& 	0.254& 	0.544& 	0.385& 	0.175& 	0.189& 	0.320& 	0.431& 	0.375   \\
          & RegNeRF\cite{niemeyer2022regnerf} & 0.568 &0.504 &0.482 &0.468 &0.445 &0.509 &0.554 &0.324 &0.268 &0.261 &0.343 &0.333 &0.218 &0.242 &0.364 &0.379 &0.389  \\
          & FreeNeRF\cite{yang2023freenerf} & 0.445 &	0.398 &	0.354 &	0.417 &	0.401 &	0.543 &	0.439 &	0.318 &	0.221 &	0.454 &	0.494 &	0.299 &	0.257 &	0.233 &	0.305 &	0.386 &	0.373  \\
          & FrugalNeRF~\cite{lin2025frugalnerf} & \cellcolor{bestcolor}0.173 & 0.590 & 0.534 & 0.506 & 0.436 & 0.553 & 0.562 & 0.539 & 0.440 & 0.554 & \cellcolor{bestcolor}0.122 & 0.573 & 0.548 & 0.369 & 0.631 & 0.552 & 0.480 \\
          & FSGS~\cite{zhu2024fsgs} & 0.264 & \cellcolor{secondcolor}0.288 & \cellcolor{secondcolor}0.281 & \cellcolor{bestcolor}0.225 & \cellcolor{secondcolor}0.214 & \cellcolor{bestcolor}0.260 & \cellcolor{thirdcolor}0.272 & \cellcolor{secondcolor}0.189 & \cellcolor{thirdcolor}0.180 & \cellcolor{thirdcolor}0.195 & 0.195 & \cellcolor{secondcolor}0.194 & 0.147 & \cellcolor{thirdcolor}0.165 & 0.322 & \cellcolor{thirdcolor}0.286 & \cellcolor{thirdcolor}0.230 \\
          & CoR-GS~\cite{zhang2024cor} & \cellcolor{thirdcolor}0.247 & \cellcolor{bestcolor}0.272 & \cellcolor{thirdcolor}0.295 & \cellcolor{secondcolor}0.226 & \cellcolor{bestcolor}0.206 & \cellcolor{secondcolor}0.264 & \cellcolor{bestcolor}0.234 & \cellcolor{thirdcolor}0.190 & \cellcolor{secondcolor}0.161 & \cellcolor{secondcolor}0.191 & \cellcolor{secondcolor}0.173 & \cellcolor{bestcolor}0.188 & \cellcolor{thirdcolor}0.145 & 0.169 & \cellcolor{secondcolor}0.211 & \cellcolor{bestcolor}0.215 & \cellcolor{bestcolor}0.212 \\
          & Ours  & 
          \cellcolor{secondcolor}0.188 & \cellcolor{thirdcolor}0.322 & \cellcolor{bestcolor}0.264 & \cellcolor{thirdcolor}0.268 & \cellcolor{thirdcolor}0.218 & \cellcolor{thirdcolor}0.339 & \cellcolor{secondcolor}0.263 & \cellcolor{bestcolor}0.188 & \cellcolor{bestcolor}0.138 & \cellcolor{bestcolor}0.170 & \cellcolor{thirdcolor}0.190 & \cellcolor{thirdcolor}0.223 & \cellcolor{bestcolor}0.110 & \cellcolor{bestcolor}0.114 & \cellcolor{bestcolor}0.170 & \cellcolor{secondcolor}0.241 & \cellcolor{secondcolor}0.213  \\

        \midrule
        \midrule
    \end{tabular}%
      \end{adjustbox}

  \label{tab:detailed_result}
  
\end{table*}%

\subsection{Comparison to Other Methods}

The quantitative and qualitative comparison results on the LEVIR-NVS dataset are presented in Table~\ref{tab:result_comparison} and Fig.~\ref{fig:vis_result}, respectively. Our approach achieves superior performance on PSNR and SSIM metrics when compared to both alternative scene representation methods and state-of-the-art few-shot NeRF approaches. Detailed performance metrics for individual scenes are provided in Table~\ref{tab:detailed_result}.

 Existing NeRF-based methods demonstrate notable limitations in few-shot rendering. NeRF and TriMipRF exhibit significant degradation issues and fail to reconstruct scene geometry accurately. ImMPI produces ghosting artifacts at image corners when multi-view inputs are insufficient, primarily due to homography transformation limitations. Although advanced few-shot NeRF variants incorporate pre-training and regularization strategies, they commonly suffer from computational inefficiency. In complex remote sensing scenes, regularization-based methods such as RegNeRF and FreeNeRF generate overly smooth results, compromising image fidelity. Despite providing semantic-level constraints, DietNeRF struggles to preserve scene details and demonstrates severe quality degradation in complex scenarios. While PixelNeRF shows promising results after scene-specific fine-tuning, its synthesized images still contain notable artifacts. As the current SOTA few-shot NeRF method, FrugalNeRF performs poorly in complex scenes—such as those containing numerous objects and significant mutual occlusion—producing noticeable noise and artifacts in the rendered images. This result suggests that the selection criterion based on reprojection error struggles to handle occlusions or scene content that is visible from only a single view. In contrast, our TriDF uses point clouds as anchors rather than dense pseudo-depth maps, enabling simpler and more reliable depth-guided optimization.

 For 3DGS-based few-shot methods, although FSGS and CoR-GS perform well overall, the proposed TriDF still demonstrates clear advantages, achieving improvements of 7.4\% in PSNR and 3.4\% in SSIM. In terms of perceptual quality, TriDF performs comparably to CoR-GS. From the rendered novel views in Fig.~\ref{fig:vis_result}, we notice that FSGS and CoR-GS exhibit visible distortions at the boundaries of the scenes. This occurs because 3DGS represents scenes explicitly through point clouds; regions not covered by the input viewpoints remain empty in the point cloud, leading to poor extrapolation ability. In contrast, implicit representations can leverage scene continuity to infer reasonable structures in adjacent regions. And our proposed method successfully addresses the typical degradation issues in few-shot rendering and achieves superior rendering quality, which accurately reconstructs intricate scene geometry. 

\begin{figure*}
    \centering
    \includegraphics[width=\textwidth]{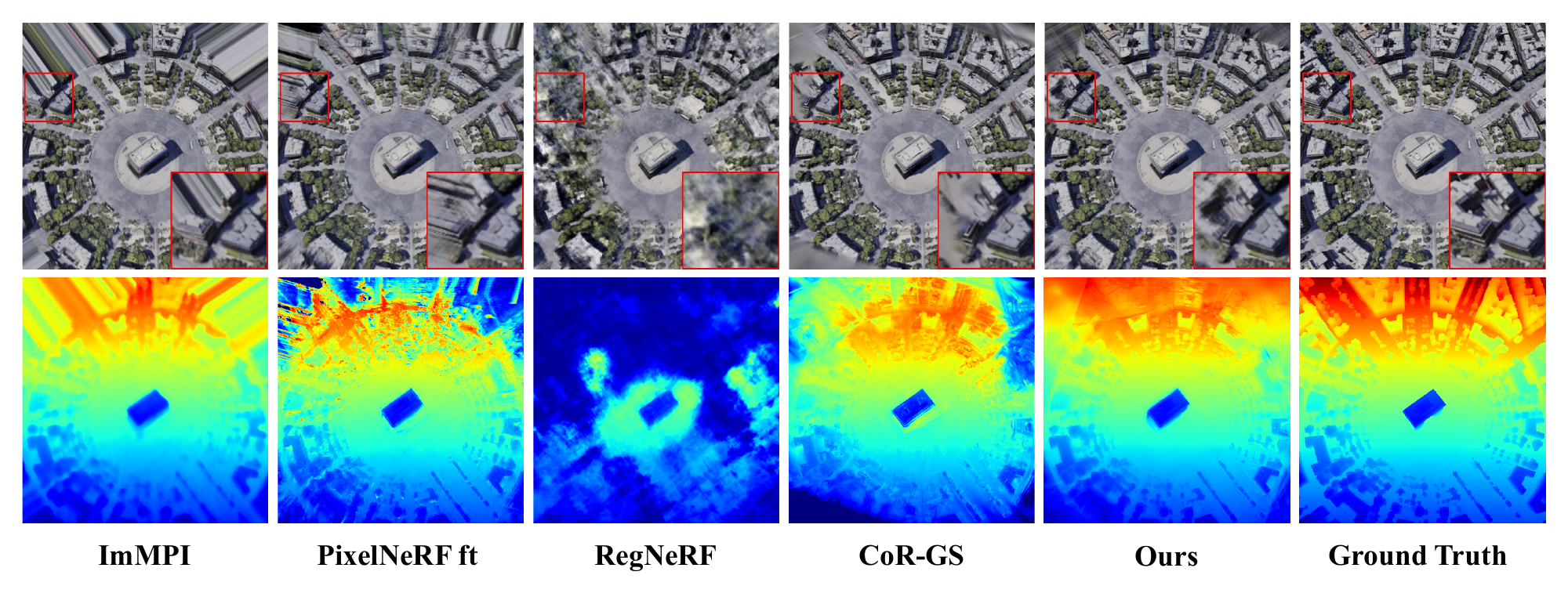}
    \caption{Detailed qualitative comparison on RGB renderings and depth maps. Our method can synthesize more accurate scene details, such as building edges and corners. The depth maps intuitively demonstrate that our approach captures scene geometry more effectively and exhibits fewer distortions.}
    \label{fig:detailed-result}
\end{figure*}

 We visualize local details and depth maps of the synthesized novel views in Fig.~\ref{fig:detailed-result}. Our method reconstructs significantly better building geometry, such as architectural edges and corners, whereas CoR-GS produces depth maps with substantial noise and discontinuities, revealing the limitations of 3DGS-based methods in accurately representing scene geometry. The other methods also exhibit varying degrees of distortion in their depth maps, making it difficult for them to accurately recover the scene geometry.

To evaluate model efficiency, we analyzed both the average training time per scene and rendering speed across different methods. As presented in Table \ref{tab:result_comparison}, our approach demonstrates competitive training efficiency. While 3DGS and DSNeRF require structure from motion for sparse point cloud estimation (a process taking merely 2 seconds with limited input views), our method necessitates additional stereo-matching and point cloud fusion steps. Nevertheless, the entire initialization process requires only 18 seconds, which is negligible compared to the overall training process. Although ImMPI exhibits competitive performance in both training efficiency and rendering quality, its rapid convergence relies on cross-scene initialization, which demands substantial pre-training time. Compared to existing state-of-the-art few-shot NeRF methods, our fully trained model reduces reconstruction time from several hours to 14 minutes. Furthermore, our early version, TriDF-10k, which utilizes only 10,000 iterations, achieves comparable rendering quality in just 5 minutes, demonstrating significant potential for real-world applications.

We also compare with 3DGS-based methods in training efficiency. FSGS is slower due to the additional regularization it introduces, while CoR-GS requires jointly training two Gaussian fields, resulting in higher memory consumption. Our method achieves a balanced trade-off between training time and memory usage. However, our rendering speed still lags far behind 3DGS-based methods designed specifically for real-time rendering.

\begin{figure*}
    \centering
    \includegraphics[width=\textwidth]{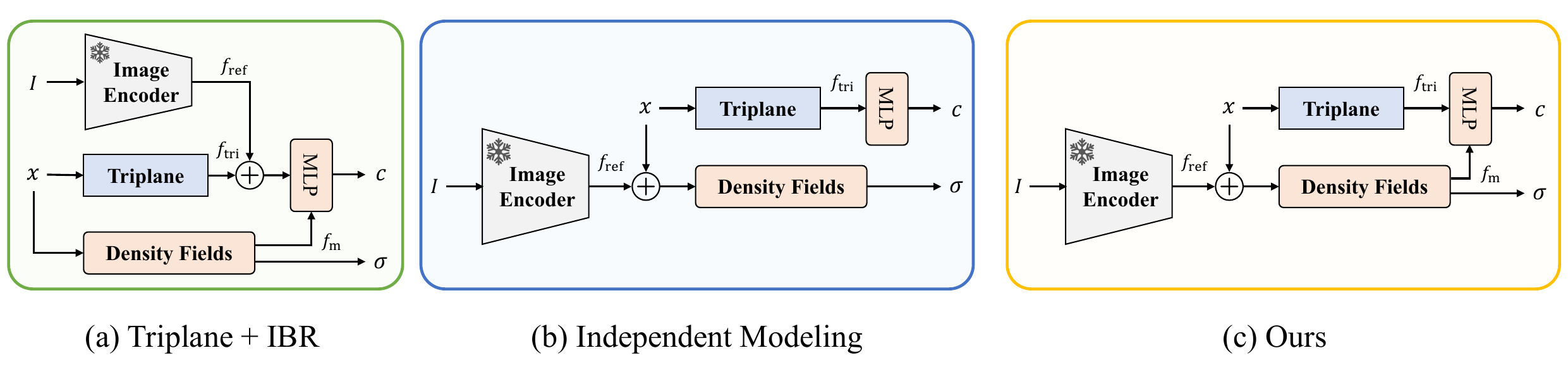}
    \caption{Three different implementations of our hybrid representation. (a) The triplane branch integrates with the image-based rendering framework, enhancing input information through feature sampling from reference images. (b) Volume density and color are modeled independently, with the density fields branch being integrated into the image-based rendering framework. (c) The density fields simultaneously generate volume density and intermediate features \( f_m \), which are subsequently processed by the triplane branch for color prediction, facilitating information interaction between color and volume density components.
    }
    \label{fig:diff-method}
\end{figure*}

\begin{figure*}
    \centering
    \includegraphics[width=\textwidth]{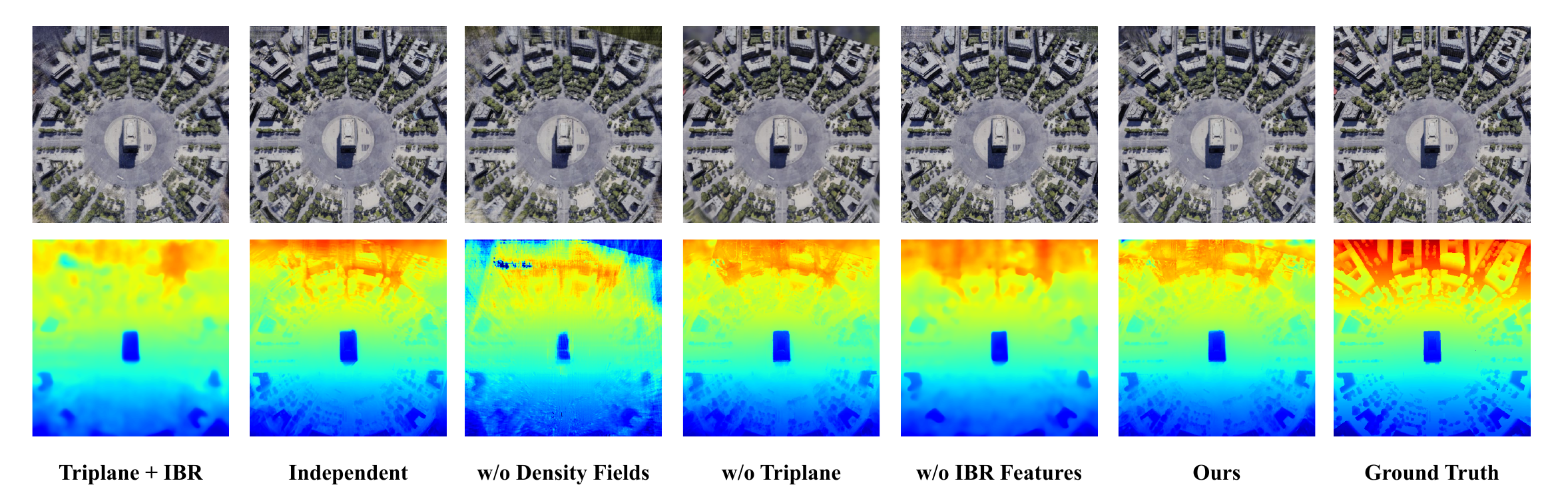}
    \caption{Visualization of rendering quality of different model structures in test view.}
    \label{fig:ablation-structure}
\end{figure*}

\subsection{Ablation Studies}

\subsubsection{Ablation of Model Structure}

We conducted comprehensive ablation experiments on our proposed hybrid representation to validate the critical design elements for efficient and accurate novel view synthesis. Our evaluation focused on three key components: the triplane representation, density fields, and the image-based rendering framework. To assess the hybrid representation thoroughly, we developed two alternative implementations for comparative analysis, as illustrated in Fig.~\ref{fig:diff-method}. These variations include the aggregation of reference features in the triplane branch and the separation of feature interactions between the triplane and density fields branches.

\begin{table}[]
    \centering
    \caption{Comparison of different model structures. 
    }
    \resizebox{0.5\textwidth}{!}{
    \begin{tabular}{l|ccc|cc}
    \toprule
     Architectures & PSNR$\uparrow$ & SSIM$\uparrow$ & LPIPS$\downarrow$ & Params$\downarrow$ & Size$\downarrow$ \\ 
    \midrule
      Triplane + IBR  &  23.10 &	0.793  & 0.238  & 8.2M & 33.0~MB\\
      Independent  &  23.48 & 0.794 & 0.234 & 8.3M & 33.3~MB\\
      Ours  &  \textbf{24.07} & \textbf{0.820} & \textbf{0.213} & 8.3M & 33.3~MB\\
      \midrule
      w/o Density Fields &  20.30 & 0.632 & 0.387 & 6.4M & 25.6~MB\\
      w/o Triplane &  23.91 & 0.809 & 0.253 & 2.0M & 8.1~MB\\
      w/o IBR Features &  22.83 & 0.773 & 0.236 & 8.2M & 32.9~MB\\
    \bottomrule
    \end{tabular}
    }
    \label{tab:structures_comparison}
\end{table}

The pre-trained encoder demonstrates the capability of capturing prior information about scene appearance and geometry. Intuitively, incorporating such priors into our scene representation (whether in the triplane or density fields) could enhance the model's spatial understanding. However, our experiments in Table \ref{tab:structures_comparison} reveal unexpected limitations when implementing the IBR framework in the triplane representation, as shown in Fig.~\ref{fig:diff-method}(a). Specifically, the triplane representation fails to learn detailed scene information due to the nature of projection transformations: the triplane features of any spatial point maintain a one-to-one correspondence with the reference image features. Given that the encoder remains frozen, the reference image features become constant. Consequently, the triplane features tend to overfit the reference features and fail to update. Based on these findings, we try to integrate IBR features into the density fields branch. Experimental results demonstrate that IBR features can provide beneficial prior information about scene structures for the density fields. In comparison to our other implementations, our method is capable of reconstructing more intricate and precise geometry, as illustrated in Fig.~\ref{fig:ablation-structure}.

\begin{table}[]
    \centering
    \caption{Effectiveness of different strategies to avoid the overfitting failure mode. 
    }
    \resizebox{0.5\textwidth}{!}{
    \begin{tabular}{l|ccc}
    \toprule
    Strategy & PSNR$\uparrow$ & SSIM$\uparrow$ & LPIPS$\downarrow$ \\
    \midrule
    None & 13.90 & 0.142 & 0.589  \\
    w/ Frequency Reg.~\cite{yang2023freenerf} & 14.40 & 0.189 & 0.578 \\
    w/ Scene Space Ann.~\cite{niemeyer2022regnerf} & \cellcolor{thirdcolor}17.62 & \cellcolor{thirdcolor}0.360 & \cellcolor{thirdcolor}0.569 \\
    w/ Sparse \({\mathcal L}_{depth}\) & \cellcolor{secondcolor}18.58 & \cellcolor{secondcolor}0.465 & \cellcolor{secondcolor}0.486 \\
    w/ Dense \({\mathcal L}_{depth}\) (Ours) & \cellcolor{bestcolor}20.08 & \cellcolor{bestcolor}0.647 & \cellcolor{bestcolor}0.339 \\
    \bottomrule
    \end{tabular}
    }
    \label{tab:optimization_comparison}
\end{table}

As illustrated in Fig.~\ref{fig:diff-method}(b), we explore modeling color and volume density as entirely independent components. While this implementation successfully generated high-quality images, as shown in Fig. \ref{fig:ablation-structure}, our final method introduces an interaction between the triplane representation and density fields. Specifically, the density fields generate supplementary features alongside volume density, which are then concatenated with triplane features for color prediction. This interactive approach demonstrates superior view synthesis quality compared to the independent modeling approach, without introducing additional computational complexity to the density fields.

\begin{figure*}
    \centering
    \includegraphics[width=\textwidth]{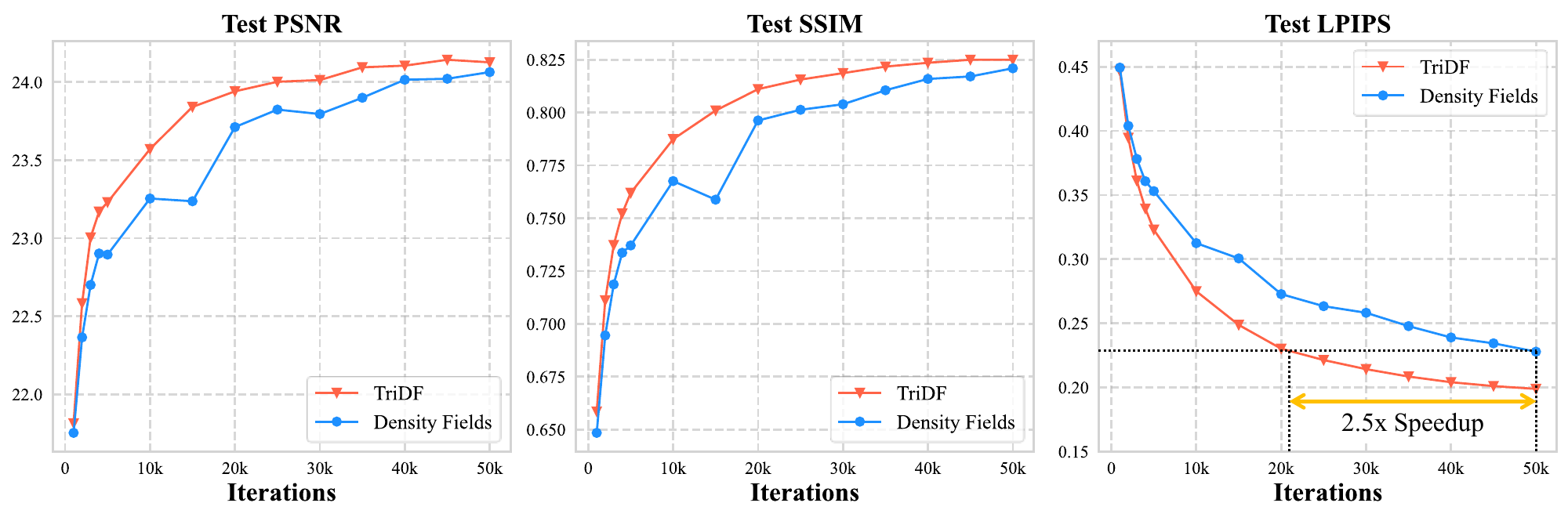}
    \caption{Comparison between TriDF and Density Fields (the model without triplane representation). We track the change in model performance within 50k iterations. }
    \label{fig:converge}
\end{figure*}

Our ablation experiments on model architecture reveal significant findings. Using the triplane representation alone proves insufficient for comprehensive scene modeling, resulting in high-frequency artifacts in synthesized images. Similarly, relying solely on the density fields branch leads to diminished perceptual quality. To quantify the triplane representation's impact on training efficiency, we compare TriDF's performance against a density-fields-only approach and track rendering quality on test views during the optimization. (Fig.~\ref{fig:converge}). The results demonstrate that incorporating the triplane representation achieves superior rendering quality with fewer iterations. Notably, TriDF requires only 20,000 iterations (40\% of the baseline) to achieve equivalent perceptual quality on LPIPS compared to the density fields approach. This enhanced efficiency can be attributed to two factors: 1) The triplane representation effectively offloads complex color information from the density fields. 2) Direct optimization of triplane features accelerates convergence. These comprehensive ablation experiments demonstrate the effectiveness of our proposed method.

\begin{table}[t]
    \centering
    \caption{Impact of Reference Feature Dimension in the Encoders on the Rendering Quality and Efficiency. 
    }
    \resizebox{0.5\textwidth}{!}{
    \begin{tabular}{c|ccc|c}
    \toprule
     Dimension & PSNR$\uparrow$ & SSIM$\uparrow$ & LPIPS$\downarrow$ & Training Time$\downarrow$\\ 
    \midrule
      3  &   20.24 & 0.771 & 0.243 & 14 min\\
      64  &  \textbf{24.07} & \textbf{0.820} & \textbf{0.213}  &  14 min\\
      128  & 21.69 & 0.659 & 0.300  & 15 min  \\
      256 &  22.71 & 0.730 & 0.261   & 16 min\\
      512 &  22.64 & 0.731 & 0.259  & 19 min  \\
    \bottomrule
    \end{tabular}
    }
    \label{tab:ref-feat-dimension}
\end{table}

\begin{table}[]
    \centering
    \caption{Performance of different Triplane and Density Fields parameters. 
    }
    \resizebox{0.5\textwidth}{!}{
    \begin{tabular}{cc|cc|ccc|cc}
    \toprule
     $N_{d}$ & $W_{d}$ &  $C_{t}$ & $R_{t}$ &  PSNR$\uparrow$ & SSIM$\uparrow$ & LPIPS$\downarrow$ & Params$\downarrow$ & Size$\downarrow$ \\ 
    \midrule
      4 & 128  & 8 & 512 &  21.38 & 0.636 & 0.396 & 6.4M & 25.8 MB\\
      4 & 256  & 8 & 512 & 22.46 & 0.707 & 0.337 & 6.6M & 26.5 MB \\
      8 & 256  & 8 & 512 & 23.38 & 0.778 & 0.258 & 6.9M & 27.6 MB \\
      8 & 512  & 8 & 512 & \textbf{24.07} & \textbf{0.820} & \textbf{0.213} & 8.3M & 33.3 MB\\
    \midrule
      8 & 512 &  8 & 256 & 23.97 & 0.805 & 0.243 & 3.6M & 14.5MB \\
      8 & 512 &  16 & 256 & 24.01 & 0.805 & 0.241 & 5.2M & 20.8MB \\
      8 & 512 &  8 & 512 & \textbf{24.07} & \textbf{0.820} & \textbf{0.213}  & 8.3M & 33.3 MB \\
      8 & 512 &  16 & 512 & 23.97 & 0.812 & 0.219 & 14.6M	& 58.5MB \\
    \bottomrule
    \end{tabular}
    }
    \label{tab:parameters}
\end{table}

\begin{figure*}
    \centering
    \includegraphics[width=\textwidth]{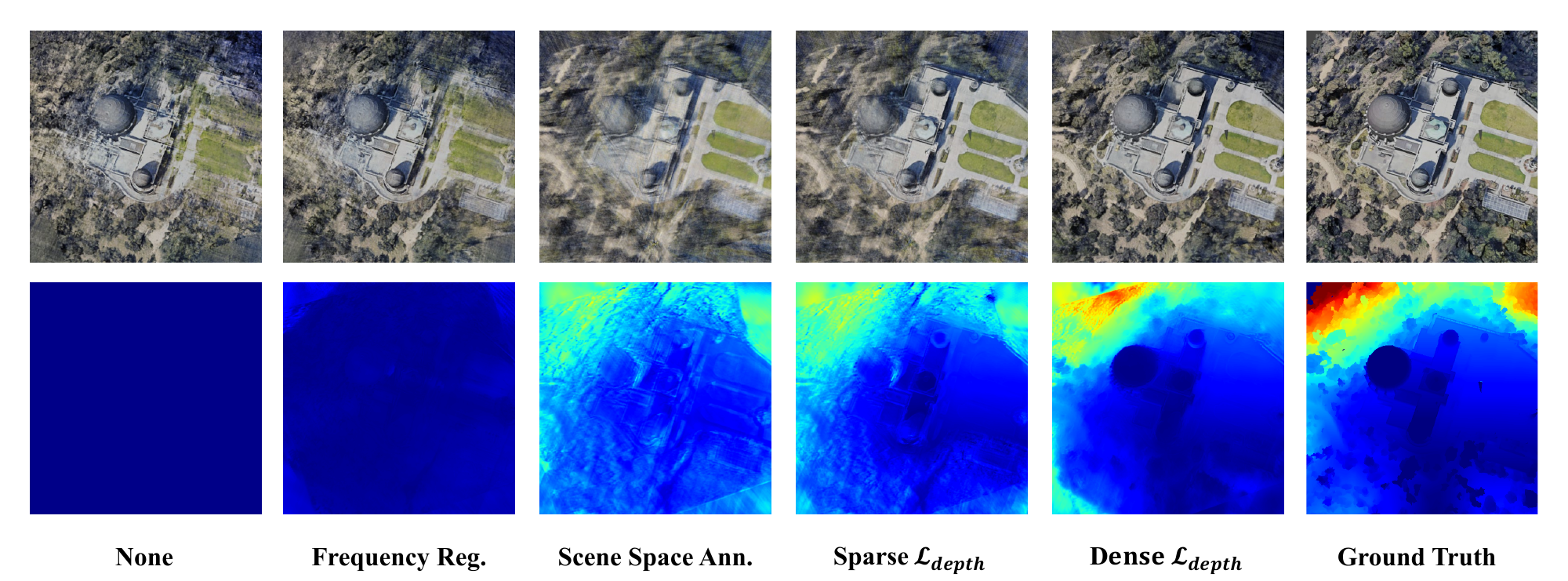}
    \caption{Comparative visualization of different optimization strategies.}
    \label{fig:vis_optimization}
\end{figure*}

\subsubsection{Ablation of Optimization Strategy}

To address typical degradation issues in few-shot rendering, we propose a depth-guided optimization approach based on point clouds. We validated our strategy's effectiveness through comparative ablation experiments as shown in Table~\ref{tab:optimization_comparison}, evaluating it against optimization strategies from advanced few-view NeRF methods, specifically the scene space annealing from RegNeRF and frequency regularization from FreeNeRF. The results demonstrate that while optimization stabilization strategies are crucial for model performance. However, existing methods' approaches perform suboptimally when applied to our hybrid representation and remote sensing scenes.

Visual comparisons of different strategies are presented in Fig.~\ref{fig:vis_optimization}. Frequency annealing proves inadequate in guiding accurate scene geometry reconstruction, while sampling space annealing fails to achieve satisfactory foreground depth reconstruction. Furthermore, we investigate the impact of the number of point clouds on depth-guided optimization by comparing sparse and dense point clouds implementations. Our method improves overall scene geometry by anchoring depth at key points. However, sparse point clouds prove insufficient to fully suppress model degradation, resulting in artifacts in synthesized images, and this issue is particularly evident in complex scenes. Increasing the point clouds number enables more accurate scene depth reconstruction and significant improvements in experimental metrics. These findings validate the necessity of augmentation based on sparse point clouds.

\begin{figure}[t]
    \centering
    \includegraphics[width=0.45\textwidth]{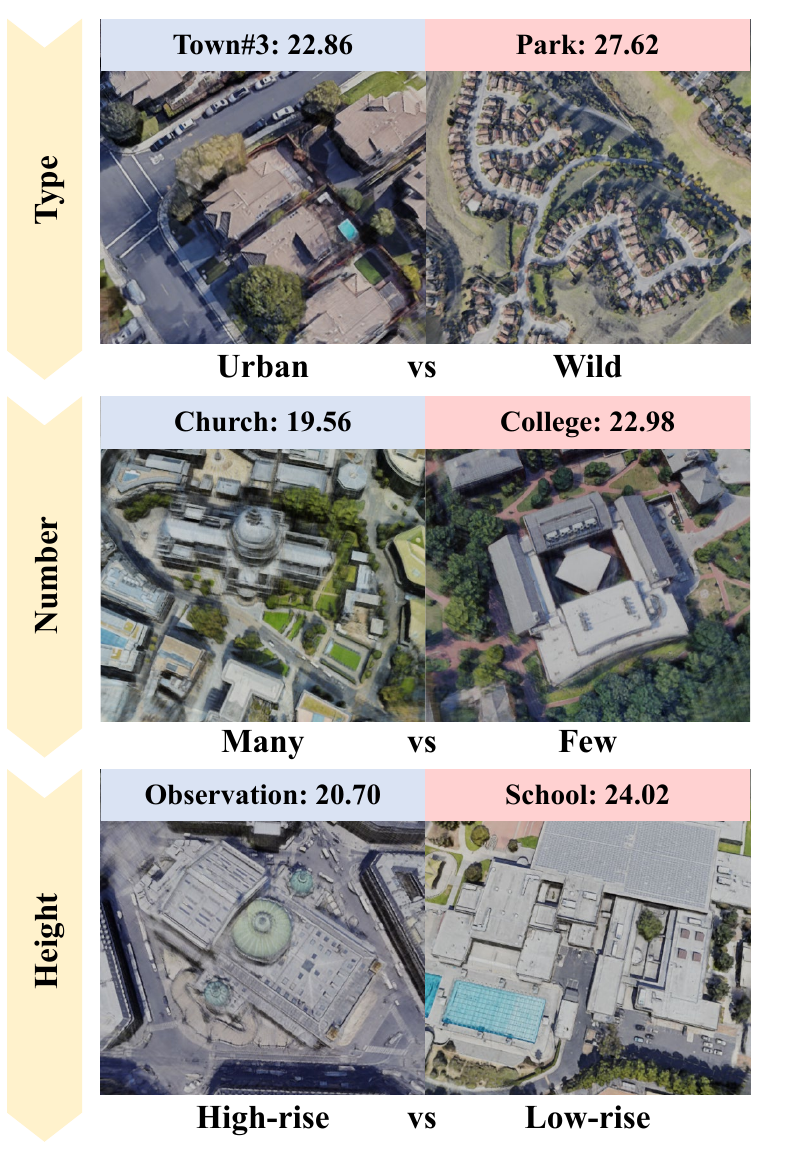}
    \caption{Comparative visualization of different scene types. We report the average PSNR on test views to evaluate the model's performance in different scenes.}
    \label{fig:vis_scene_type}
\end{figure}

\subsubsection{Impacts of Model Parameters}

Model parameter configuration significantly influences rendering quality and training efficiency. We first investigate the impact of reference feature dimensions, as shown in Table~\ref{tab:ref-feat-dimension}. A dimension of 3 represents raw RGB values as reference features, while dimensions from 64 to 512 represent aggregated outputs from various layers of the pre-trained encoder. Our experiments reveal that the model performs optimally when aggregating shallow-layer features with a dimension of 64. Model performance degrades and training time increases with higher feature dimensions.

Further experiments investigate the effects of density fields and triplane representation parameters (Table~\ref{tab:parameters}). For the density fields, we assess model performance using different combinations of network depth \(N_d\) (4, 8) and width \(W_d\) (256, 512). Results indicate that reducing the density fields size significantly diminishes model performance, suggesting that insufficient model capacity fails to capture detailed scene structures.

For the triplane representation, we evaluate the impact of spatial resolution and feature dimensions using combinations of plane resolutions \(R_t\) (256, 512) and feature dimensions \(C_t\) (8, 16). Our findings demonstrate that triplane resolution is the primary performance determinant. Low resolution significantly reduces sampling precision, preventing the model from capturing distinguishable high-frequency color features and resulting in degraded perceptual quality. Conversely, high feature dimensionality introduces unnecessary redundancy. Optimal performance is achieved with \(C_t=8\) and \(R_t=512\).

Additionally, we observe that the explicit triplane representation accounts for most parameters in the hybrid representation, with model size increasing rapidly at higher triplane resolutions. Notably, merely reducing the size of density fields or triplane does not effectively enhance training efficiency.

\subsubsection{Performance on Different Scenes}

We also evaluate the model’s performance across different types of scenes. As illustrated in the 
Fig.~\ref{fig:vis_scene_type}, we roughly categorize the influencing factors into three parts: ground object types, the number of buildings, and building height. First, scenes with high natural vegetation coverage (e.g., Mountain and Park) exhibit better reconstruction quality than scenes with dense building coverage. This is because natural vegetation tends to have simpler and more tolerant texture structures, making them less sensitive to errors. In contrast, buildings contain fine-grained geometric details, where even small deviations can lead to a noticeable degradation in synthesis quality. The number of buildings directly reflects the complexity of a scene. Scenes with fewer buildings (e.g., Building \#1 and College) provide more complete view coverage, resulting in higher-quality synthesis. Conversely, scenes with many buildings (such as Church) are more difficult to reconstruct due to their intricate structures and the occlusions between buildings. Finally, due to the limited number of viewpoints and restricted shooting angles, taller buildings introduce more severe occlusion. As a result, scenes with relatively flat terrain (e.g., Factory and School) tend to achieve better reconstruction quality.

\section{Discussion}

TriDF achieves a balance between quality and efficiency in few-shot NVS in remote sensing scenes, distinguishing itself from existing methods. Our hybrid representation demonstrates superior rendering quality compared to fast reconstruction methods while maintaining a significant efficiency advantage over few-shot NeRF approaches. This success highlights the potential of hybrid representations in novel view synthesis. The integration of explicit representations' fast inference capabilities with implicit representations' continuous geometry modeling enables the development of more efficient, higher-quality synthesis methods. Furthermore, leveraging the inherent sparsity of 3D content through compact representations, such as triplane, holds great promise for future research.

However, our method has several limitations. First, it struggles to reconstruct regions with limited viewpoint coverage, including areas visible from only a single view or completely occluded regions. These areas suffer from image blur and inaccurate depth estimation due to inherent ambiguities. Second, the depth-guided optimization depends on estimated point clouds, potentially leading to degraded performance in scenes too complex to generate sufficient point cloud data. Future research should focus on addressing these challenges.

\section{Conclusion}
In this work, we explore the potential of triplane representation in remote sensing novel view synthesis and propose an efficient hybrid 3D representation, Triplane-Accelerated Density Fields  (TriDF). Our approach achieves superior rendering quality from sparse viewpoint inputs while achieving a faster reconstruction speed compared to NeRF-based methods. We improve the training efficiency by decoupling color and volume density, leveraging the triplane representation to model high-frequency color information while utilizing the density fields to capture continuous volume density. Furthermore, we introduce an efficient depth-guided optimization technique to prevent overfitting issues in few-shot NVS. Experiments conducted across multiple scenes demonstrate that our approach outperforms other state-of-the-art few-shot NVS methods while maintaining fast reconstruction within only 5 minutes. Consequently, TriDF exhibits immense potential for 3D interpretation in remote sensing applications.


%

\bibliographystyle{IEEEtran}
\bibliography{reference}

\end{document}